\documentclass{article}

    \PassOptionsToPackage{numbers, compress}{natbib}

\usepackage[preprint]{neurips_2026}

\usepackage[utf8]{inputenc} 
\usepackage[T1]{fontenc}    
\usepackage{hyperref}       
\usepackage{url}            
\usepackage{booktabs}       
\usepackage{amsfonts}       
\usepackage{nicefrac}       
\usepackage{microtype}      
\usepackage{xcolor}         

\usepackage{multirow}
\usepackage{adjustbox}
\usepackage{tcolorbox}
\usepackage{listings}
\usepackage{xcolor}
\usepackage{wrapfig}
\usepackage{booktabs}
\usepackage{colortbl}
\usepackage{subcaption}

\lstdefinestyle{plainins}{
    backgroundcolor=\color{white},   
    commentstyle=\color{codegreen},
    keywordstyle=\color{magenta},
    numberstyle=\tiny\color{codegray},
    stringstyle=\color{codepurple},
    basicstyle=\ttfamily\small,
    breakatwhitespace=false,         
    breaklines=true,                 
    captionpos=b,                    
    keepspaces=true,                 
    numbers=none,                    
    numbersep=5pt,                  
    showspaces=false,                
    showstringspaces=false,
    showtabs=false,                  
    tabsize=2,
    aboveskip=0pt,
    belowskip=0pt,
    frame=single,
    escapeinside={(*@}{@*)}
}

\newcommand{\rparagraph}[1]{\vspace{1.2mm}\noindent\textbf{#1.}}

\definecolor{clrImgSearch}{HTML}{60A5FA}
\definecolor{clrTxtSearch}{HTML}{F87171}

\newcommand{\imgsearch}[1][]{\textcolor{clrImgSearch}{<\texttt{image\_search}>#1}}                                      
\newcommand{\txtsearch}[1][]{\textcolor{clrTxtSearch}{<\texttt{text\_search}>#1}}  
\usepackage{amsmath}
\usepackage{makecell}
\usepackage{multirow}
\usepackage{booktabs}
\usepackage{colortbl}
\usepackage{multirow}
\usepackage{algorithm}
\usepackage{algpseudocode}
\usepackage{enumitem}
\usepackage{xcolor}
\usepackage{alltt}
\usepackage{tcolorbox}
\usepackage{adjustbox}
\makeatletter
\def\blfootnote{\gdef\@thefnmark{}\@footnotetext}
\makeatother
\tcbuselibrary{skins, breakable}

\definecolor{ourblue}{RGB}{208,228,246}
\definecolor{deltagreen}{RGB}{34,139,34}
\title{HyperEyes: Dual-Grained Efficiency-Aware Reinforcement Learning for Parallel Multimodal Search Agents}

\definecolor{ourrow}{HTML}{E8F5E9}       
\definecolor{toolblue}{HTML}{0037C9}    
\definecolor{groupgray}{HTML}{F1F3F4}   
\definecolor{deltagreen}{HTML}{2E7D32}  
\definecolor{basecolor}{RGB}{230, 236, 245}   
\definecolor{ours1}{RGB}{240, 246, 240}       
\definecolor{ours2}{RGB}{225, 238, 225}       
\definecolor{ours3}{RGB}{210, 230, 210}       
\definecolor{lightbluebg}{RGB}{235, 245, 255}  
\definecolor{blueframe}{RGB}{70, 130, 180} 
\usepackage{fvextra}
\DefineVerbatimEnvironment{VerbatimWrap}{Verbatim}{
breaklines=true, 
breakindent=0pt, 
breaksymbol={},
fontsize=\scriptsize,
}

\author{
    Guankai Li\textsuperscript{1,\,$\dagger$,\,$\ddagger$} \quad
    Jiabin Chen\textsuperscript{1,\,$\dagger$} \quad
    Yi Xu\textsuperscript{2} \quad
    Xichen Zhang\textsuperscript{1} \quad
    Yuan Lu\textsuperscript{1,\,$\ast$} \\[1em]
    \textsuperscript{1}Xiaohongshu Inc. \\
    \textsuperscript{2}University of Cambridge
}

\begin{document}

\blfootnote{{\normalfont $^\dagger$Equal Contributor.\;\;$^\ddagger$Project Lead.\;\;$^\ast$Corresponding author: Yuan Lu (\texttt{luyuan2@xiaohongshu.com})}}

\maketitle
\vspace{-2em}
\begin{abstract}

Existing multimodal search agents process target entities sequentially, issuing one tool call per entity and accumulating redundant interaction rounds whenever a query naturally decomposes into independent sub-retrievals. For such decomposable queries, we argue that effective multimodal agents should search \emph{wider} rather than \emph{longer}: dispatching multiple grounded queries concurrently within a round, rather than sequentially. To this end, we present HyperEyes, a parallel multimodal search agent that fuses visual grounding and retrieval into a single atomic action, enabling concurrent search across multiple entities while treating inference efficiency as a first-class training objective. HyperEyes is trained in two stages: for cold-start supervision, we develop a \textbf{Parallel-Amenable Data Synthesis Pipeline} covering visual multi-entity and textual multi constraint queries, and curate efficiency-oriented trajectories via Progressive Rejection Sampling. Building on this foundation, our central contribution, a \textbf{Dual-Grained Efficiency-Aware Reinforcement Learning} framework, operates at two complementary levels. At the macro level, we propose TRACE (Tool-use Reference-Adaptive Cost Efficiency), a trajectory-level reward whose reference is monotonically tightened during training to suppress superfluous tool calls without over-restricting genuine multi-hop search. At the micro level, we adapt On-Policy Distillation (OPD) to the multimodal agentic search setting, injecting dense token-level corrective signals from an external teacher on failed rollouts to mitigate the credit-assignment deficiency of sparse outcome rewards. Since most existing multimodal search benchmarks evaluate accuracy as the sole metric, omitting inference cost and parallel-search capability, we further introduce IMEB, a human-curated benchmark that jointly evaluates multimodal search capability and efficiency, comprising 300
multi-entity visual instances. Across six benchmarks, HyperEyes-30B surpasses the strongest open-source multimodal search agent of comparable scale by \textbf{9.9\%} in accuracy with \textbf{5.3$\times$ fewer} tool-call rounds on average. Code \& Data are publicly available at \url{https://github.com/DeepExperience/HyperEyes}.

\end{abstract}

\section{Introduction}

\begin{figure}[!ht]
    \centering
    \includegraphics[width=1.0\linewidth]{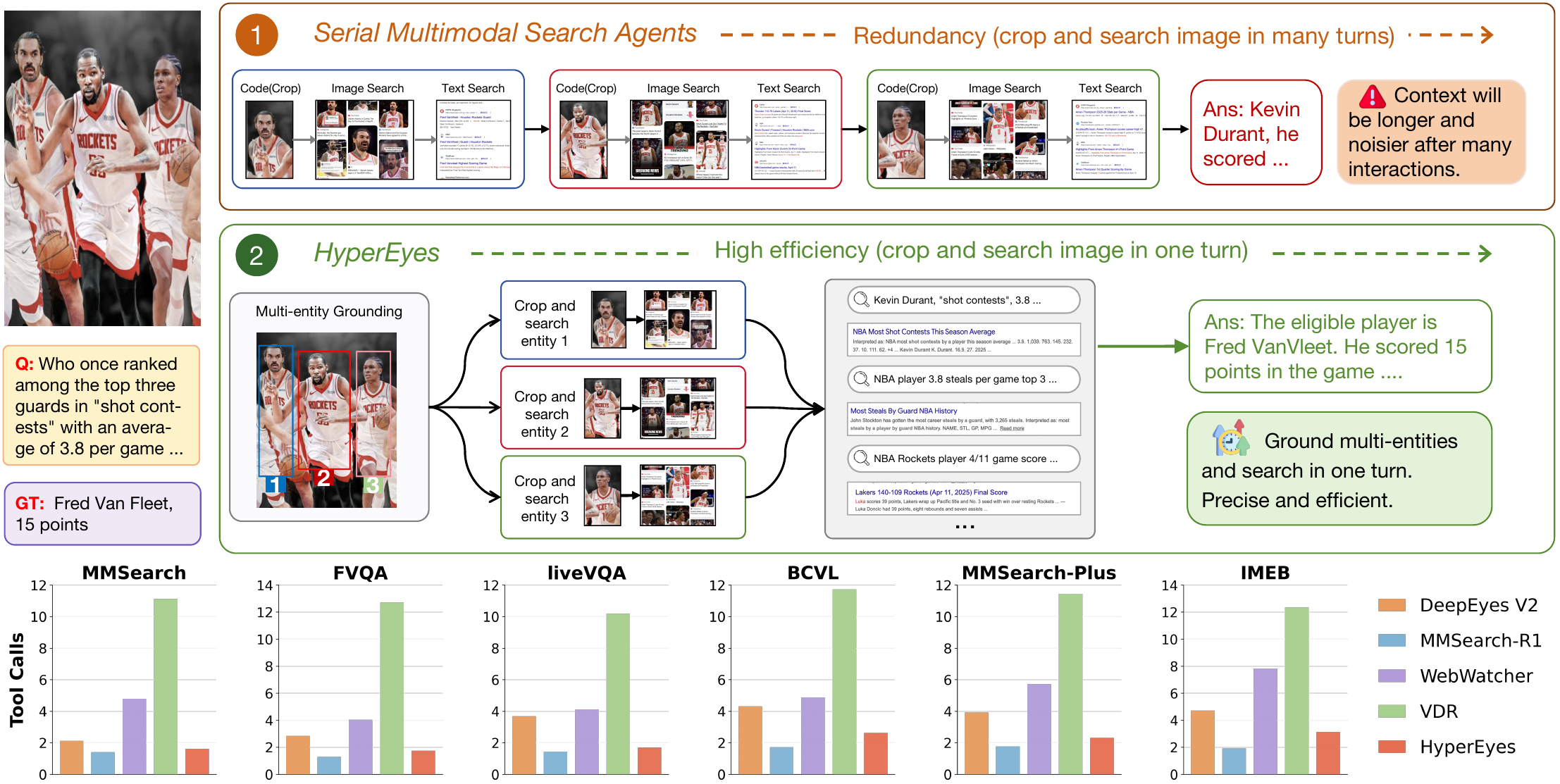}
    \caption{Comparison between conventional multimodal search agents and HyperEyes. While conventional agents suffer from redundant interaction rounds to process multiple entities, HyperEyes achieves high efficiency by grounding and searching multiple entities concurrently in a single turn.}
    \label{fig:motivation}
\end{figure}
The parametric knowledge of Large Language Models (LLMs)~\cite{brown2020language, ouyang2022training, anil2023palm} and Multimodal Large Language Models (MLLMs)~\cite{gemini2024, hurst2024gpt} is structurally constrained by their training data cutoff. This limitation drives the development of search agents~\cite{yao2023react, jin2025searchr}, which actively invoke external retrieval tools to ground responses in real-time, verifiable information. However, the prevailing paradigm of multimodal search agents relies heavily on sequential tool invocations to deepen the reasoning chain~\cite{hong2026deepeyesv, wu2025mmsearch, geng2025webwatcher, chu2026redsearcher}. While effective for multi-hop reasoning tasks, this sequential approach incurs severe interaction redundancy when queries can be decomposed into independent sub-retrievals. 

Although parallel tool invocation has emerged in text-based agents~\cite{zhao2025parallelsearch, lin2026w, ko2026hybrid} and recent visual models~\cite{huang2026vision} to address this bottleneck, possessing parallel capability does not guarantee efficient search behavior. As existing models~\cite{geng2025webwatcher, hong2026deepeyesv, wu2025mmsearch} are optimized primarily through pure accuracy rewards, they lack the incentive to prefer a compact parallel trajectory over a verbose one. Consequently, without explicit efficiency objectives, parallel capability often degrades into brute-force over-searching, forcing models to undergo numerous unnecessary interaction rounds to recover accuracy.

To overcome this fundamental inefficiency, we propose \textbf{HyperEyes}, a parallel multimodal search agent designed around the principle of ``search wider, not longer.'' As illustrated in Figure~\ref{fig:motivation}, whereas conventional agents suffer from redundant interaction rounds to process multiple entities, HyperEyes achieves high efficiency by grounding and searching multiple entities concurrently in a single turn. It operates on a Unified Grounded Search (UGS) action space that fuses visual grounding and retrieval into a single atomic action, extending text-level parallelism to the visual modality. To ensure the learned policy is parallel and strictly non-redundant, we pair this architecture with a \textbf{Dual-Grained Efficiency-Aware} reinforcement learning (RL) framework that treats efficiency as a primary optimization objective. At the macro level, it features TRACE, a trajectory-level reference that dynamically tightens during training to guide the policy toward optimal efficiency. At the micro level, it introduces On-Policy Distillation (OPD) \citep{gu2024minillm}, which resolves ambiguous credit assignment by providing dense per-token supervision from an expert teacher on failed rollouts. Furthermore, we support this training paradigm with a Parallel-Amenable Data Synthesis Pipeline, which utilizes Progressive Rejection Sampling to curate high-quality, efficiency-oriented cold-start trajectories.

Standard evaluations~\cite{jiang2024mmsearch, tao2025mmsearch, fu2025seeking, geng2025webwatcher}, however, primarily assess final answer accuracy, masking the inefficiencies of verbose search trajectories. To quantify the efficiency gains achieved by parallel search, we introduce the Image Multi-Entity Benchmark (\textbf{IMEB}), a human-curated dataset that pioneers the joint evaluation of multimodal search agents on both accuracy and search efficiency. Each instance features a multi-entity image paired with a question that strictly requires concurrent localization and retrieval across multiple entities. Under this comprehensive evaluation, we demonstrate that parallel search breadth acts as the primary bottleneck in multi-entity visual search.

In summary, our main contributions are as follows:

\begin{itemize}
    \item \textbf{Parallel multimodal search agent.} We propose HyperEyes, an efficient agent operating on a Unified Grounded Search action space. We optimize it via a Parallel-Amenable Data Synthesis pipeline and a Dual-Grained Efficiency-Aware RL framework, combining dynamic trajectory-level efficiency constraints with token-level On-Policy Distillation.
    
    \item \textbf{Efficiency-aware benchmark.} We introduce IMEB, the first human-curated benchmark to jointly evaluate answer accuracy and search efficiency, establishing operational efficiency as a first-class metric in multi-entity visual scenarios.
    
    \item \textbf{Strong empirical performance.} Across six benchmarks, HyperEyes-30B establishes state-of-the-art results. It Pareto-dominates existing models, surpassing the strongest open-source agent by 9.9\% in accuracy while requiring 5.3$\times$ fewer tool-call rounds on average.
\end{itemize}

\section{Related Work}

\subsection{Text-based Search Agents}

To overcome the inherent limitations of static, single-hop Retrieval-Augmented Generation (RAG)~\citep{lewis2020retrieval} in resolving complex, multi-hop queries, information seeking has fundamentally shifted toward Agentic Deep Research. While early frameworks bridged this gap via iterative prompting (e.g., ReAct \citep{yao2023react}, Self-Ask \citep{press-etal-2023-measuring}) and supervised fine-tuning, the current frontier focuses intensely on long-horizon search and robust multi-turn tool calling to tackle sophisticated, open-domain challenges. Search-R1 \citep{jin2025searchr} treats web navigation as a sequential decision-making process optimized via Reinforcement Learning (RL). Advanced frameworks such as DeepDive~\citep{lu2025deepdive} and MiroThinker \citep{team2025mirothinker} push the boundaries of complex multi-step planning, enabling models to track dynamic states, execute iterative tool invocations, and maintain goal consistency over extended reasoning cycles. Concurrently, the rise of Test-Time Scaling (TTS) \citep{zhang2025survey} has catalyzed deep investigations into the optimal allocation of inference computation. Recent paradigms exploring mechanisms like "wide search" and the "search more, think less" strategy systematically evaluate the trade-offs between expansive external knowledge gathering and deep internal reasoning, demonstrating that scaling exploratory search steps can effectively alleviate the cognitive burden on the LLM's reasoning engine. However, despite these algorithmic leaps in long-horizon planning, pure text search agents remain fundamentally bottlenecked by their unimodal nature. When navigating the real-world web, they inevitably suffer from critical semantic loss upon encountering visually rich evidence—such as data charts, spatial UI layouts, or explicitly image-grounded constraints. This modality constraint highlights an urgent imperative to transcend text-only boundaries, naturally paving the way for unified multimodal search agents capable of holistic visual-semantic reasoning.

\subsection{Multi-modality Search Agents}

Multimodal Large Language Models (MLLMs) have rapidly evolved from passive perception engines into agentic systems capable of actively interacting with dynamic environments. Current research predominantly focuses on empowering MLLMs with long-horizon search capabilities and multi-tool orchestration to tackle complex, knowledge-intensive queries, as evaluated by recent multi-hop benchmarks like FVQA \citep{wu2025mmsearch}, MMSearch-Plus \citep{tao2025mmsearch}, and BrowseComp-VL (BC-VL) \citep{geng2025webwatcher}. To navigate these challenges, recent frameworks have actively embraced the "Think-Act-Observe" paradigm. For instance, DeepMMSearch-R1 \citep{narayan2025deepmmsearch} and DeepEyesV2 \citep{hong2026deepeyesv} introduce "thinking with images" by executing active visual manipulations (e.g., cropping, rotating, or marking via generated code) to extract fine-grained features before initiating web retrieval. Meanwhile, agents like WebWatcher \citep{geng2025webwatcher} and Skywork-R1V4 \citep{zhang2025skywork} integrate diverse tools (e.g., code interpreters, text/image search) through Reinforcement Learning (RL) or high-fidelity supervised fine-tuning to facilitate in-depth information seeking. Taking a broader approach, Vision-DeepResearch (VDR) \citep{huang2026vision} tackles hit-rate issues in noisy web environments by formalizing a multi-turn, multi-scale trial-and-error retrieval paradigm, significantly pushing the boundaries of long-horizon multimodal planning.

Despite these remarkable leaps in orchestrating long-horizon reasoning, existing multimodal search agents still suffer from compounded inefficiencies, particularly in multi-entity scenarios. First, processing multiple visual entities often induces sequential tool invocations, leading to prohibitive end-to-end latency that is further exacerbated by the initialization overhead of code-execution sandboxes. Second, decoupled "manipulate-then-search" paradigms are inherently brittle, as early visual localization errors can irreversibly cascade into downstream retrieval and reasoning failures. Finally, current training strategies predominantly supervise final answer correctness without penalizing redundant tool usage, inadvertently incentivizing an "over-retrieval" behavior that inflates token consumption and introduces distracting noise into the context. Consequently, mitigating these fundamental bottlenecks to achieve efficient, parallelized, and redundancy-aware multimodal search remains a critical unresolved challenge.

\section{HyperEyes}
\label{sec:hypereyes}

\subsection{Formulation}
\label{sec:formulation}  

Following the ReAct paradigm~\citep{yao2023react}, HyperEyes operates as an iterative reasoning-and-acting agent. Given a query $q$, the agent produces a trajectory
\begin{equation}
\tau = \bigl(q,\; (r_0, a_0, o_0),\; (r_1, a_1, o_1),\; \ldots,\; (r_T, a_T, o_T),\; y\bigr),
\end{equation}
where at each turn $t$, the agent $\pi_\theta$ generates a reasoning trace $r_t$ over the accumulated context $h_t = \bigl(q,\,(r_0, a_0,o_0),\ldots,(r_{t-1}, a_{t-1},o_{t-1})\bigr)$, selects a tool call $a_t \sim \pi_\theta(\cdot \mid h_t, r_t)$, and receives an observation $o_t$ from the retrieval environment. This process iterates until the agent provides a final answer $y$ or reaches the maximum allowed turn $T$.

\paragraph{Multimodal search tools.} 
To enable interaction with the real-world internet, the agent is equipped with two tools: (i) Image search, invoked via \imgsearch, retrieves visually relevant results for a grounded visual image. (ii) Text search, invoked via \txtsearch, retrieves textual evidence given a natural language query. 

\paragraph{Unified grounded search.}
Existing agents adopt a two-stage ``crop-then-search'' pipeline~\citep{hong2026deepeyesv}, introducing brittle dependencies where an early localization error corrupts downstream search results. Furthermore, this separation forecloses parallelism, as the agent must wait for each image crop to be produced, forcing multi-entity queries into sequential chains. We address this with Unified Grounded Search (UGS), reformulating visual grounding from a prerequisite step into a parameter of the retrieval action. By simultaneously predicting bounding boxes for all target entities, UGS allows the policy to dispatch parallel search queries across modalities within a single turn (see Appendix~\ref{Prompt Tempaltes}).

\begin{figure}[!ht]
\centering
\includegraphics[width=1.1\linewidth]{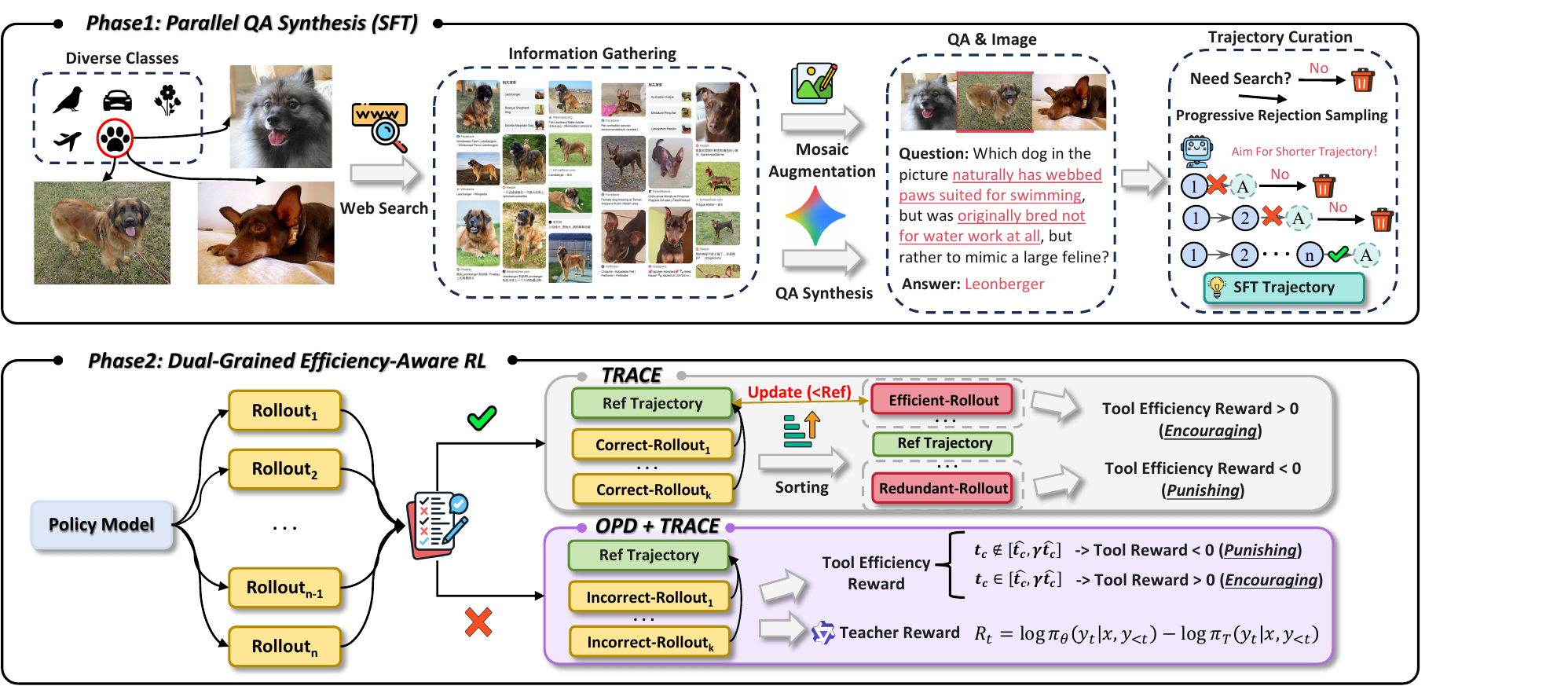}
\caption{Overview of the HyperEyes training framework. The framework consists of two main phases: (1) a Parallel-Amenable Data Synthesis pipeline that constructs multi-entity QA pairs and curates efficient trajectories, and (2) a Dual-Grained Efficiency-Aware RL algorithm that optimizes parallel search behavior through trajectory-level efficiency rewards and token-level distillation.}
\label{fig:framework}
\end{figure}

\subsection{Training Data Curation}
\label{sec:met_Data}

Current multimodal corpora predominantly feature single-entity or chain-style reasoning, lacking queries that explicitly demand parallel tool invocation. To establish robust cold-start supervision and enable efficiency-aware optimization, we design a comprehensive three-stage data curation pipeline (illustrated in Fig.~\ref{fig:framework}). First, we compile a diverse pool of tasks by aggregating public datasets and synthesizing novel multi-entity queries (Sec.~\ref{subsubsec:qa}). Second, we construct a high-quality Supervised Fine-Tuning (SFT) dataset using Progressive Rejection Sampling to distill parallel, non-redundant trajectories (Sec.~\ref{subsubsec:traj}). Third, we isolate medium-difficulty samples to build a specialized Reinforcement Learning (RL) dataset (Sec.~\ref{subsubsec:rl_data}). The overall data composition is detailed in Table~\ref{tab:data_composition}. We defer comprehensive algorithmic details to Appendix~\ref{appendix:data-curation}.

\begin{table}[t]
\small
\centering
\caption{Composition of the training dataset.}
\label{tab:data_composition}
\vspace{2mm}
\renewcommand{\arraystretch}{0.95} 
\resizebox{0.95\linewidth}{!}{%
\begin{tabular}{l r r r r}
\toprule
\textbf{Data Source} & \textbf{\# QA Pairs} & \textbf{\# SFT} & \textbf{\# 30B RL} & \textbf{\# 235B RL} \\
\midrule

\rowcolor{groupgray}
\multicolumn{5}{l}{\textit{Public Benchmarks}} \\
\quad LiveVQA \citep{fu2025seeking}          & 100k & 13.5k & 3k   & 5k   \\
\quad REDSearch \citep{chu2026redsearcher}   & 10k  & 2k    & 0.5k & 0.5k \\
\quad InfoSeek \citep{xia2025open}           & 41k  & 3k    & --   & --   \\
\quad iNaturalist \citep{vendrow2024inquire} & 75k  & 2.5k  & 2k   & 3k   \\
\quad Google-Landmark \citep{49052}          & 12k  & 1k    & --   & --   \\
\quad DeepDive \citep{lu2025deepdive}        & 3k   & 1.5k  & --   & --   \\

\rowcolor{groupgray}
\multicolumn{5}{l}{\textit{Ours}} \\
\quad Internal Human Annotations                                   & 5k   & 0.5k  & --   & --   \\
\quad Visual Multi-Entity (Sec.\ref{subsubsec:qa})      & 20k  & 5k    & 0.5k & 0.5k \\
\quad Textual Multi-Constraint (Sec.\ref{subsubsec:qa}) & 5k   & 1k    & --   & --   \\
\midrule

\rowcolor{ourrow}
\textbf{Total} & \textbf{271k} & \textbf{30k} & \textbf{6k} & \textbf{9k} \\
\bottomrule
\end{tabular}%
}
\end{table}
\subsubsection{Task Formulation and Synthesis}
\label{subsubsec:qa}

We compile a rich foundation of 246,000 multi-hop reasoning and visual recognition queries from existing public benchmarks and internal human annotations. To strictly enforce parallel search behaviors, we supplement this pool with 25,000 novel synthetic queries across two bespoke pipelines.

\paragraph{Visual multi-entity synthesis.}
As shown in the data synthesis pipeline of Fig.~\ref{fig:framework}, we start with a collection of fine-grained visual classification datasets~\cite{marsili2025same}. For each class, a knowledge retriever gathers structured attribute knowledge to build a per-class knowledge base. Images from distinct classes are then sampled and composited via mosaic augmentation into multi-entity scenes. Conditioned on the knowledge base, a question synthesizer generates QA pairs that require integrating retrieved information across all co-occurring entities. Consequently, omitting any single entity precludes the model from deducing the correct answer. This pipeline yields 20,000 visual multi-entity QA pairs.

\paragraph{Textual multi-constraint synthesis.}
Deviating from conventional chain-style reasoning, we construct queries demanding answers that satisfy multiple independent attribute constraints. Using Wikidata~\cite{vrandevcic2014wikidata} as the source, we perform a multi-hop random walk to collect candidate entities. From the attributes of these candidates, we sample $m\ge2$ predicates whose intersection defines the unique ground-truth set. This textual pipeline contributes an additional 5,000 complex queries.

\paragraph{Tool-necessity filtering.}
We apply a unified filter across all task sources, systematically discarding any QA pair that Qwen3-VL-235B~\cite{bai2025qwen3} successfully resolves without external tool access, thereby finalizing our foundational pool of 271,000 genuinely tool-dependent tasks.

\subsubsection{SFT Trajectory Curation}
\label{subsubsec:traj}

\paragraph{Progressive rejection sampling.}
Naive agentic rollouts often suffer from redundant tool calls and iterative query reformulations, which inflate latency without improving correctness. To obtain a clean, efficiency-oriented training signal, we propose Progressive Rejection Sampling (PRS), depicted in the trajectory curation module of Fig.~\ref{fig:framework}. Taking the 271,000 initial queries as input, PRS samples trajectories across an ascending schedule of turn budgets, strictly retaining the shortest successful trajectory for each query (Algorithm~\ref{alg:prs}). Because restrictive budgets inherently preclude iterative refinement, the surviving trajectories naturally exhibit single-turn precision and parallel execution.

\paragraph{Quality filtering.}
Relying solely on outcome correctness is insufficient, as successful trajectories might entail parametric guessing or uninformative actions. We further discard trajectories exhibiting format invalidity, zero information gain, or ungrounded reasoning. Through this cascade of sampling and quality filtering, the initial pool of 271,000 tasks is distilled to 30,000 high-fidelity trajectories, ensuring the SFT dataset instills optimal, zero-redundancy parallel dispatch behaviors.

\subsubsection{RL Data Selection}
\label{subsubsec:rl_data}

To support sequence-level optimization in the subsequent reinforcement learning phase, we curate specialized subsets of medium-difficulty queries from the PRS pipeline. Specifically, we isolate 6,056 and 9,337 queries for the 30B and 235B models, respectively, where the initial model fails to find an answer under the tightest pass@1 setting but successfully resolves the task under relaxed pass@5 constraints. The initial successful trajectories from these selected samples establish vital dynamic efficiency boundaries for the RL reward mechanism.

\subsection{Agentic Training}
\label{sec:met_training}

To elicit and refine the parallel tool-use capabilities of HyperEyes, we employ a two-stage agentic training paradigm. We first fine-tune the model on the curated demonstration corpus to instill basic parallel retrieval behaviors. Subsequently, we apply a Dual-Grained Efficiency-Aware RL framework to optimize search efficiency and token-level credit assignment.

\subsubsection{Supervised Fine-Tuning}
\label{subsec:mt_sft}

The Supervised Fine-Tuning (SFT) phase optimizes the base MLLM via next-token prediction on the curated trajectory corpus (Sec.~\ref{sec:met_Data}). Because these trajectories undergo strict efficiency filtering, the SFT policy directly internalizes one-shot parallel dispatch without learning to iteratively reformulate queries. However, pure behavior cloning lacks sequence-level optimization for end-to-end inference efficiency, necessitating a dedicated reinforcement learning intervention.

\subsubsection{Reinforcement Learning}
\label{subsec:GRPO}

The SFT policy inherits two critical limitations. First, it lacks explicit optimization for inference efficiency, often resulting in redundant tool invocations. Second, sparse outcome-based rewards fail to provide fine-grained supervision to isolate reasoning errors during complex parallel planning. To resolve these issues, we propose a Dual-Grained Efficiency-Aware RL framework. At the macro level, we employ Group Relative Policy Optimization (GRPO) \citep{shao2024deepseekmathpushinglimitsmathematical} with a novel Reference-Adaptive Cost Efficiency (TRACE) reward to explicitly optimize tool-use efficiency. At the micro level, On-Policy Distillation (OPD) leverages a strong teacher model $\pi_{\text{teacher}}$ to inject dense token-level corrective signals exclusively into failed trajectories for the student model $\pi_{\theta}$ (Fig.~\ref{fig:framework}).

\paragraph{TRACE: Tool-use reference-adaptive cost efficiency.}
The core challenge of rewarding efficiency lies in determining a ``reasonable'' number of tool calls, which is inherently query-dependent. A static threshold proves either too loose to suppress redundancy or too tight to accommodate legitimate multi-hop searches. TRACE addresses this by providing an evolvable efficiency reference. The total reward for a trajectory is formulated as:
\begin{equation}
R = R_{\text{acc}} + R_{\text{fmt}} + R_{\text{tool}},
\end{equation}
where $R_{\text{acc}} \in \{0,1\}$ acts as a binary correctness judge, $R_{\text{fmt}} \in \{0, -\lambda_{\text{fmt}}\}$ penalizes schema parsing failures, and $R_{\text{tool}}$ serves as the core adaptive efficiency reward.

We characterize the tool usage of a trajectory by two dimensions: the number of tool-call rounds $t_c$ and the total number of tool invocations across all rounds $t_s$. For each medium-difficulty sample in the RL dataset (Sec.~\ref{subsubsec:rl_data}), the values of $t_c$ and $t_s$ from its initial successful trajectory serve as the initial references $\hat{t}_c^{(0)}$ and $\hat{t}_s^{(0)}$. During training, the primary round reference $\hat{t}_c$ tightens per epoch:
\begin{equation}
\hat{t}_c^{(e+1)} = \min\left(\hat{t}_c^{(e)},\, t_c^{(e+1)}\right), \quad e = 0, 1, \ldots, n-1,
\end{equation}
where $t_c^{(e+1)}$ represents the minimum $t_c$ among successful rollouts during epoch $e+1$. This update rule guarantees a monotonically tightening reference threshold, forming an implicit curriculum that anchors the reward boundary at a level just attainable by the current policy. The total invocation reference $\hat{t}_s$ simultaneously updates to mirror the tool consumption of that minimal-round trajectory.

The TRACE reward is then defined as:
\begin{equation}
R_{\text{tool}} = 
\begin{cases}
-\lambda_{\text{red}}, & R_{\text{acc}} = 0 \text{ and } \left(t_c < \hat{t}_c \text{ or } t_c > \gamma \hat{t}_c\right) \\
0, & R_{\text{acc}} = 0 \text{ and } t_c \in [\hat{t}_c,\, \gamma \hat{t}_c] \\
R^+ \in [r_{\text{min}}^+, r_{\text{max}}^+], & R_{\text{acc}} = 1,\, t_c > 0,\, t_c \leq \hat{t}_c \text{ and } t_s \leq \hat{t}_s \\
R^- \in [r_{\text{min}}^-, r_{\text{max}}^-], & R_{\text{acc}} = 1,\, t_c > 0,\, \text{otherwise},
\end{cases}
\end{equation}
where $\gamma > 1$ acts as a redundancy tolerance factor, and $\lambda_{\text{red}}$ applies a constant penalty. To provide continuous optimization signals within discrete bounds, $R^+$ and $R^-$ undergo linear interpolation based on intra-group rank. For a sampled group of size $G$, let $\rho \in \{1, \dots, G\}$ denote the ascending rank of a trajectory's $t_c$ (where $\rho=1$ is the most efficient). The assigned reward scales dynamically as:
\begin{equation}
R^* = r_{\text{min}}^* + \frac{G - \rho}{G - 1} \left( r_{\text{max}}^* - r_{\text{min}}^* \right), \quad \text{for } * \in \{+, -\}.
\end{equation}
Crucially, trajectories receive positive rewards only when falling in the strictly efficient region ($t_c \leq \hat{t}_c$ and $t_s \leq \hat{t}_s$). Incorporating the $t_s$ constraint elegantly prevents reward hacking, a scenario where the model minimizes interaction rounds by exhaustively spamming parallel calls within a single turn. Furthermore, correct trajectories with $t_c = 0$ receive $R_{\text{tool}} = 0$ to prevent parametric guessing. Finally, these aggregated rewards $R_i$ normalize within the group to compute the relative advantage $\hat{A}_i = (R_i - \mu_R)/\sigma_R$ for the GRPO objective.

\paragraph{Token-level Correction on Failed Rollouts.}
Because TRACE operates at the trajectory level, it exhibits a credit-assignment deficiency on failed rollouts (i.e.,$R_{\text{acc}}=0$): a uniform negative advantage indiscriminately penalizes every token, including valid intermediate reasoning steps and accurately issued tool calls that precede the final mistake. To recover learning signals from these correct intermediate steps, OPD distills token-level supervision from a frozen teacher $\pi_{\text{teacher}}$ into the student $\pi_{\theta}$, applied exclusively to failed rollouts. Specifically, we minimize the reverse KL divergence over completion tokens. Its mode-seeking nature drives the student to concentrate on the high-probability reasoning modes of the teacher rather than averaging over them. Confining the KL term strictly to failed rollouts ensures that the parallel-dispatch behaviors discovered by TRACE on successful rollouts remain completely untouched. Combing this, the final student loss is defined as:
\begin{equation}
\mathcal{L}(\theta) = \mathcal{L}_{\text{GRPO}}(\theta)
+ \lambda_{\text{kd}}\,\mathbb{E}_{\tau\sim\pi_{\theta}^{\text{old}}}\!\left[
\mathbf{1}[R_{\text{acc}}(\tau)=0]\cdot
\tfrac{1}{|\mathcal{A}_\tau|}\!\sum_{t\in\mathcal{A}_\tau}\!
\mathrm{KL}\!\bigl(\pi_\theta(\cdot\!\mid\!s_t)\,\|\,\pi_{\text{teacher}}(\cdot\!\mid\!s_t)\bigr)\right],
\end{equation}
where the rollouts $\tau$ are shared with GRPO, $\mathcal{A}_\tau$ represents the set of completion tokens in $\tau$, and $\lambda_{\text{kd}}$ scales the distillation strength. The teacher remains frozen and $\pi_{\theta}^{\text{old}}$ acts as a sampling-only reference; gradients flow only through $\theta$. Under this design, TRACE shapes successful exploration at the trajectory level while OPD provides dense correction at the token level, allowing the student to absorb the reasoning patterns of the teacher without inheriting its inference cost.

\begin{figure}[!ht]
    \centering
    \includegraphics[width=1\linewidth]{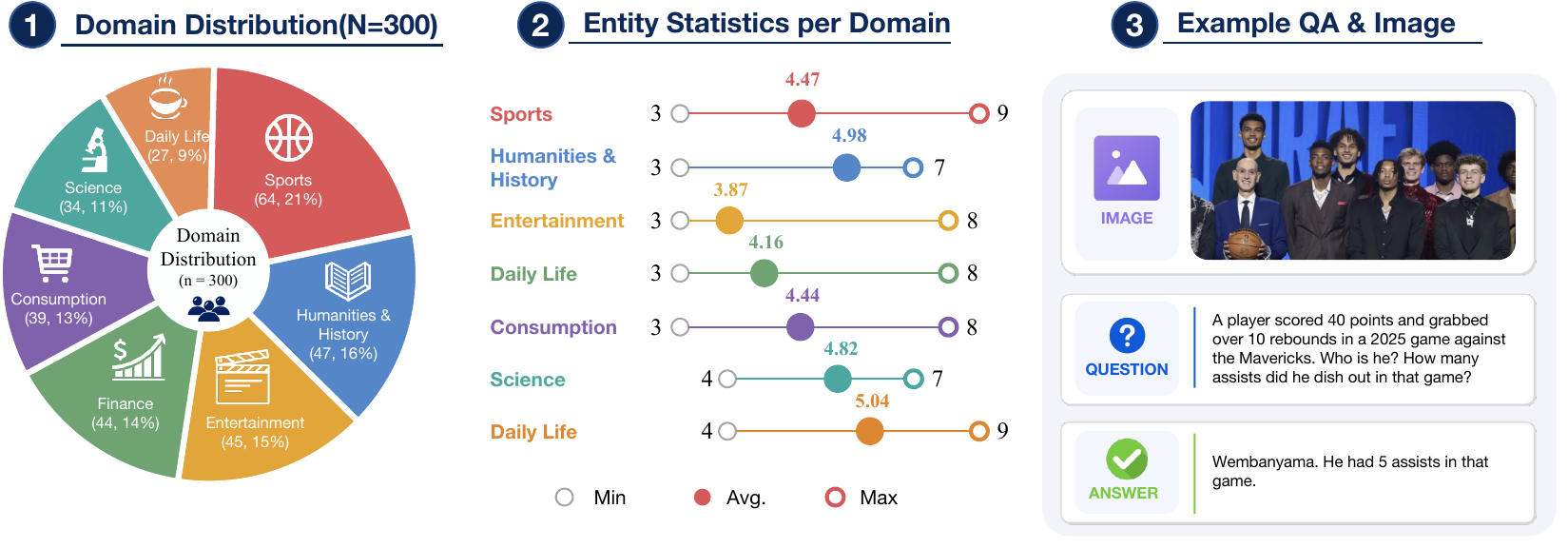}
    \caption{Overview of the IMEB benchmark, including domain distribution ($N=300$), entity count statistics for each domain, and an example question-answer pair.}
    \label{fig:IMEB}
\end{figure}

\section{Construction of IMEB Benchmark}
\label{sec:met_IMEB}

Existing multimodal search benchmarks evaluate reasoning accuracy while neglecting tool-call efficiency~\cite{jiang2024mmsearch,wu2025mmsearch,geng2025webwatcher,tao2025mmsearch}. Consequently, models resolve parallelizable queries via sequential trajectories, inflating latency and introducing noisy retrievals. The Image Multi-Entity Benchmark~(IMEB) addresses this gap by elevating search efficiency to a primary evaluation axis, constructing queries that require concurrent localization and retrieval across multiple entities.

Curated by PhD-level annotators through multiple rounds of double-blind cross-validation, IMEB comprises 300 rigorously verified instances across diverse domains, with an average of 4.6 entities per image~(Figure~\ref{fig:IMEB}). Every question undergoes rigorous human peer-review and automated filtering to guarantee that it is unambiguously solvable yet strictly necessitates concurrent external tool invocation. We defer the comprehensive curation pipeline to Appendix~\ref{appendix:IMEB-data}. Since traditional accuracy metrics alone cannot capture parallel operational efficiency, we further propose a unified metric.

\paragraph{Cost-aware score.}
To jointly quantify reasoning correctness and search efficiency, we introduce the Cost-Aware Score~(CAS):
\begin{equation} 
\label{eq:cas} 
\mathrm{CAS} = \frac{\mathrm{Acc}^2 \times 100}{N_{\mathrm{tok}} + 2 N_{\mathrm{tool}} + 1}.
\end{equation}
The squared accuracy term ensures correctness remains the primary optimization objective. The denominator penalizes token consumption~($N_{\mathrm{tok}}$, in thousands) and sequential tool-call rounds~($N_{\mathrm{tool}}$). These weights approximate a one-second latency overhead for both generation and tool execution, facilitating fair comparisons across distinct agent architectures.

\section{Experiment}
\label{sec:exp}

\begin{table*}[t]
\centering
\small
\caption{Main results (accuracy \%) on six multimodal search benchmarks. \textbf{Bold} = best, \underline{underline} = second-best. $\Delta$ rows show absolute improvement of HyperEyes over the second-best open-source model under the Agentic Workflow setting. "–" denotes unreported results. For brevity, we abbreviate HyperEyes as HE in the table. Accuracy numbers are taken from the original papers, while tool-call turns and metrics missing from the original papers are obtained via local deployment and inference of their open-source models.}
\setlength{\tabcolsep}{6pt} 
\renewcommand{\arraystretch}{1.3} 

\begin{adjustbox}{width=\textwidth}
\begin{tabular}{l | cccccc | c}
\toprule
\textbf{Model} & \textbf{MMSearch} & \textbf{FVQA} & \textbf{LiveVQA} & \textbf{BCVL} & \textbf{MMSearch+} & \textbf{IMEB} & \textbf{Avg.} \\
\midrule

\rowcolor{groupgray}
\multicolumn{8}{c}{\textit{Direct Answer}} \\
Qwen3-VL-30B         & 21.3 / \textcolor{toolblue}{--}  & 36.7 / \textcolor{toolblue}{--} & 35.6 / \textcolor{toolblue}{--} & 17.2 / \textcolor{toolblue}{--} & 2.1 / \textcolor{toolblue}{--}  & 6.7 / \textcolor{toolblue}{--}  & 19.8 / \textcolor{toolblue}{--} \\
Qwen3-VL-235B        & 30.3 / \textcolor{toolblue}{--} & 44.2 / \textcolor{toolblue}{--} & 41.4 / \textcolor{toolblue}{--} & 21.8 / \textcolor{toolblue}{--} & 6.9 / \textcolor{toolblue}{--}  & 12.0 / \textcolor{toolblue}{--} & 26.1 / \textcolor{toolblue}{--} \\
Kimi-K2.5            & 65.6 / \textcolor{toolblue}{--} & 59.6 / \textcolor{toolblue}{--} & 57.3 / \textcolor{toolblue}{--} & 27.6 / \textcolor{toolblue}{--} & 9.7 / \textcolor{toolblue}{--}  & 27.7 / \textcolor{toolblue}{--} & 41.2 / \textcolor{toolblue}{--} \\
Claude-Opus-4.6                & 59.8 / \textcolor{toolblue}{--} & 60.1 / \textcolor{toolblue}{--} & 53.1 / \textcolor{toolblue}{--} & 43.5 / \textcolor{toolblue}{--} & 13.2 / \textcolor{toolblue}{--} & 27.0 / \textcolor{toolblue}{--} & 42.8 / \textcolor{toolblue}{--} \\
Gemini-3.1-Pro       & 75.4 / \textcolor{toolblue}{--} & 62.7 / \textcolor{toolblue}{--} & 51.5 / \textcolor{toolblue}{--} & 53.1 / \textcolor{toolblue}{--} & 21.0 / \textcolor{toolblue}{--} & 40.8 / \textcolor{toolblue}{--} & 50.7 / \textcolor{toolblue}{--} \\
\midrule

\rowcolor{groupgray}
\multicolumn{8}{c}{\textit{Agentic Workflow}} \\
Qwen3-VL-30B         & 54.1 / \textcolor{toolblue}{1.7} & 58.0 / \textcolor{toolblue}{2.0} & 49.8 / \textcolor{toolblue}{1.9} & 29.0 / \textcolor{toolblue}{4.4} & 9.7 / \textcolor{toolblue}{2.8}  & 17.7 / \textcolor{toolblue}{4.3}  & 36.4 / \textcolor{toolblue}{2.7} \\
Qwen3-VL-235B        & 64.8 / \textcolor{toolblue}{1.4} & 70.2 / \textcolor{toolblue}{1.7} & 58.2 / \textcolor{toolblue}{1.6} & 37.9 / \textcolor{toolblue}{2.7} & 20.3 / \textcolor{toolblue}{4.0} & 30.0 / \textcolor{toolblue}{4.8} & 46.9 / \textcolor{toolblue}{2.7} \\
Kimi-K2.5            & 76.6 / \textcolor{toolblue}{2.2} & 76.5 / \textcolor{toolblue}{2.5} & 76.6 / \textcolor{toolblue}{2.1} & 50.3 / \textcolor{toolblue}{5.1} & 27.8 / \textcolor{toolblue}{3.1} & \textbf{55.3} / \textcolor{toolblue}{8.8} & 60.5 / \textcolor{toolblue}{4.0} \\

Claude-Opus-4.6     & 76.2 / \textcolor{toolblue}{1.6} & 74.5 / \textcolor{toolblue}{1.3} & 67.4 / \textcolor{toolblue}{1.2} & 48.3 / \textcolor{toolblue}{2.4} & 31.3 / \textcolor{toolblue}{2.4} & 41.7 / \textcolor{toolblue}{3.4} & 56.5 / \textcolor{toolblue}{2.0} \\

Gemini-3.1-Pro       & 86.1 / \textcolor{toolblue}{1.2} & \textbf{84.0} / \textcolor{toolblue}{1.3} & 76.6 / \textcolor{toolblue}{1.4} & \textbf{64.1} / \textcolor{toolblue}{2.0} & \textbf{44.2} / \textcolor{toolblue}{2.9} & 51.3 / \textcolor{toolblue}{2.1} & \textbf{67.7} / \textcolor{toolblue}{1.8} \\
\midrule

\rowcolor{groupgray}
\multicolumn{8}{c}{\textit{Multimodal Deep Search Agents}} \\
DeepEyes-V2          & 63.7 / \textcolor{toolblue}{2.1} & 60.6 / \textcolor{toolblue}{2.8} & 58.0 / \textcolor{toolblue}{3.7} & 24.8 / \textcolor{toolblue}{4.3} & 9.5 / \textcolor{toolblue}{3.9}  & 18.0 / \textcolor{toolblue}{4.7} & 39.1 / \textcolor{toolblue}{3.6} \\
MMSearch-R1          & 53.8 / \textcolor{toolblue}{1.4} & 58.4 / \textcolor{toolblue}{1.3} & 48.4 / \textcolor{toolblue}{1.4} & 19.1 / \textcolor{toolblue}{1.7} & 10.1 / \textcolor{toolblue}{1.8} & 3.3 / \textcolor{toolblue}{1.9}  & 32.2 / \textcolor{toolblue}{1.6} \\
WebWatcher           & 55.3 / \textcolor{toolblue}{4.8} & 64.3 / \textcolor{toolblue}{4.0} & 58.7 / \textcolor{toolblue}{4.1} & 27.0 / \textcolor{toolblue}{4.9} & 11.5 / \textcolor{toolblue}{5.7} & 15.3 / \textcolor{toolblue}{7.8} & 38.7 / \textcolor{toolblue}{5.2} \\
VDR                  & 69.6 / \textcolor{toolblue}{11.1}& 74.2 / \textcolor{toolblue}{12.7}& 77.6 / \textcolor{toolblue}{10.2}& 53.7 / \textcolor{toolblue}{11.7}& 28.5 / \textcolor{toolblue}{11.4}& 21.2 / \textcolor{toolblue}{12.3}& 54.1 / \textcolor{toolblue}{11.6} \\
REDSearch            & 72.9 / \textcolor{toolblue}{--}  & -- / \textcolor{toolblue}{--}   & 79.3 / \textcolor{toolblue}{--} & 57.2 / \textcolor{toolblue}{--} & 26.6 / \textcolor{toolblue}{--} & -- / \textcolor{toolblue}{--}   & -- / \textcolor{toolblue}{--} \\
\midrule

\rowcolor{groupgray}
\multicolumn{8}{c}{\textit{Ours}} \\
HE-30B (SFT)         & 82.0 / \textcolor{toolblue}{1.8} & 76.1 / \textcolor{toolblue}{2.0} & 80.3 / \textcolor{toolblue}{1.9} & 47.6 / \textcolor{toolblue}{3.9} & 25.0 / \textcolor{toolblue}{3.7} & 42.0 / \textcolor{toolblue}{3.8} & 58.8 / \textcolor{toolblue}{2.9} \\
HE-30B (RL)         & \underline{86.9} / \textcolor{toolblue}{1.6} & 79.3 / \textcolor{toolblue}{1.7} & 81.6 / \textcolor{toolblue}{1.7} & 57.9 / \textcolor{toolblue}{2.6} & 31.5 / \textcolor{toolblue}{2.3} & 46.7 / \textcolor{toolblue}{3.1} & 64.0 / \textcolor{toolblue}{2.2} \\
\rowcolor{ourrow}
\textcolor{deltagreen}{\textbf{$\Delta$}} & \textcolor{deltagreen}{\textbf{+14.0 / -9.5}} & \textcolor{deltagreen}{\textbf{+5.1 / -11.0}} & \textcolor{deltagreen}{\textbf{+2.3 / -8.5}} & \textcolor{deltagreen}{\textbf{+0.7 / -9.1}} & \textcolor{deltagreen}{\textbf{+3.0 / -9.1}} & \textcolor{deltagreen}{\textbf{+25.5 / -9.2}} & \textcolor{deltagreen}{\textbf{+9.9 / -9.4}} \\
HE-235B (SFT)        & 84.4 / \textcolor{toolblue}{1.7} & 80.3 / \textcolor{toolblue}{1.9} & \underline{83.7} / \textcolor{toolblue}{2.1} & 54.4 / \textcolor{toolblue}{3.7} & 31.8 / \textcolor{toolblue}{3.9} & 50.0 / \textcolor{toolblue}{3.3} & 64.1 / \textcolor{toolblue}{2.8} \\
HE-235B (RL)         & \textbf{88.5} / \textcolor{toolblue}{1.4} & \underline{81.4} / \textcolor{toolblue}{1.5} & \textbf{84.1} / \textcolor{toolblue}{1.5} & \underline{60.0} / \textcolor{toolblue}{2.2} & \underline{32.6} / \textcolor{toolblue}{2.2} & \underline{52.7} / \textcolor{toolblue}{3.0} & \underline{66.6} / \textcolor{toolblue}{2.0} \\
\rowcolor{ourrow}
\textcolor{deltagreen}{\textbf{$\Delta$}} & \textcolor{deltagreen}{\textbf{+15.6 / -9.7}} & \textcolor{deltagreen}{\textbf{+7.2 / -11.2}} & \textcolor{deltagreen}{\textbf{+4.8 / -8.7}} & \textcolor{deltagreen}{\textbf{+2.8 / -9.5}} & \textcolor{deltagreen}{\textbf{+4.1 / -9.2}} & \textcolor{deltagreen}{\textbf{+31.5 / -9.3}} & \textcolor{deltagreen}{\textbf{+12.5 / -9.6}} \\
\bottomrule
\end{tabular}
\end{adjustbox}
\label{tab:main_results}
\end{table*}
\subsection{Experimental Setup}
\label{subsub:setup}

\paragraph{Implementation.} We instantiate HyperEyes on two backbone models, Qwen3-VL-30B and Qwen3-VL-235B \citep{bai2025qwen3}. We first conduct a cold start for the models using 30,000 curated trajectories. During the RL phase, we select a subset of medium-difficulty samples from the parallel QA corpus and optimize the policy via GRPO with TRACE. For the 30B variant, we additionally enable OPD, designating HyperEyes-235B as the teacher model to supply dense token-level guidance. Full training details and hyperparameters are provided in Appendix~\ref{appendix:training}.

\paragraph{Baselines.} We compare HyperEyes against three groups of baselines. The first group consists of the native MLLM backbones Qwen3-VL-30B-A3B-Instruct and Qwen3-VL-235B-A22B-Instruct. The second group includes representative multimodal search agents, namely DeepEyes-V2~\cite{hong2026deepeyesv}, MMSearch-R1~\cite{wu2025mmsearch}, WebWatcher~\citep{geng2025webwatcher}, VDR~\cite{huang2026vision}, and REDSearch~\cite{chu2026redsearcher}. The third group covers leading commercial models, including Kimi-K2.5~\cite{team2026kimi}, Claude-Opus-4.6 \citep{anthropic2026opus}, and Gemini-3.1-Pro \citep{gemini2026pro}.

\paragraph{Benchmarks and Metrics.} We evaluate HyperEyes on six multimodal search benchmarks that span complementary scenarios: MMSearch~\cite{jiang2024mmsearch}, FVQA~\cite{wu2025mmsearch} and LiveVQA~\cite{fu2025seeking} for shallow-hop visual search; BrowseComp-VL (BCVL)~\cite{geng2025webwatcher} and MMSearch-Plus~\cite{tao2025mmsearch} for complex multi-hop visual reasoning; and our newly proposed IMEB, the first benchmark designed to quantify efficiency in multi-entity grounded retrieval. To comprehensively assess performance, we report two metrics: Accuracy (Acc), judged by an LLM-as-a-judge against ground-truth answers, and Average Tool-Call Turns (Turns), which measures the number of decoder forward passes that emit a tool-call block.

\subsection{Main Results}

We evaluate HyperEyes against open-source agents and proprietary frontier models across six multimodal search benchmarks (Table~\ref{tab:main_results}), alongside cost-accuracy trade-offs on BCVL and IMEB utilizing the proposed CAS metric (Table~\ref{tab:CAS}). Three key observations emerge from this evaluation.

\paragraph{Open-source state-of-the-art on the accuracy and efficiency Pareto frontier.}
HyperEyes establishes a new standard for open-source search agents. As Table~\ref{tab:main_results} demonstrates, HyperEyes-235B~(RL) approaches the proprietary Gemini-3.1-Pro while substantially outperforming Claude-Opus-4.6 and Kimi-K2.5. The marginal performance gap between HyperEyes and Gemini primarily originates from the richer parametric knowledge of the latter, evidenced by its superior ``direct answer'' capabilities. Notably, the compact HyperEyes-30B~(RL) exceeds the leading open-source search agent of comparable scale, VDR~\cite{huang2026vision}, by $9.9$ absolute accuracy points. More crucially, HyperEyes Pareto dominates existing models in both accuracy and operational efficiency. For instance, compared to VDR~\cite{huang2026vision}, HyperEyes-30B achieves superior accuracy while requiring $5.3\times$ fewer tool calls. A deeper comparison reveals that conventional serial agents suffer from redundant crop-then-search loops, whereas HyperEyes issues unified grounded searches that decouple entity identification from downstream reasoning, thereby drastically reducing latency and cascading errors.

\paragraph{RL jointly pushes the accuracy ceiling and shrinks tool-call redundancy.}
During the SFT stage, HyperEyes already surpasses all open-source baselines by leveraging high-quality trajectories derived from Progressive Rejection Sampling. The subsequent RL stage advances both accuracy and efficiency simultaneously. For example, HyperEyes-235B improves its average accuracy to $66.6\%$ while reducing the tool-call turns on complex benchmarks such as BCVL. This dual improvement confirms that TRACE's adaptive efficiency reference and OPD's token-level signal provide complementary supervision. Furthermore, as our analysis in Appendix~\ref{appendix:more_calls} reveals, blindly increasing tool calls frequently degrades final-answer accuracy due to the accumulation of distractor evidence. By explicitly penalizing over-retrieval, our RL framework guides the policy to fuse multiple sources holistically, rendering it highly robust against noisy retrieval contexts (further detailed in Appendix~\ref{sec:robustness}).

\paragraph{Best cost-aware accuracy under the CAS metric.}
Traditional accuracy metrics fail to penalize the verbose search trajectories common in multi-entity scenarios. Under the proposed CAS metric, which jointly evaluates accuracy and inference cost, HyperEyes demonstrates a commanding advantage. As Table~\ref{tab:CAS} indicates, HyperEyes-30B achieves the highest CAS on both complex multi-hop (BCVL) and multi-entity (IMEB) benchmarks, exceeding the second-best open-source competitors by massive margins of $4.3\times$ and $7.6\times$, respectively. Rather than trading cost for accuracy, HyperEyes delivers substantially higher information density per unit of compute. As illustrated by the comparative case study in the supplementary material, replacing serial isolation with parallel multi-entity grounding eliminates intermediate noise accumulation, translating to swift and highly accurate task resolution.

\begin{table*}[t]
\centering
\caption{Comparison of different methods on \textbf{BCVL} and \textbf{IMEB} benchmarks.}
\label{tab:CAS}
\setlength{\tabcolsep}{6pt}
\renewcommand{\arraystretch}{1.25}
\definecolor{ourgray}{RGB}{230,238,247}
\definecolor{rulecolor}{RGB}{60,60,60}
\arrayrulecolor{rulecolor}

\resizebox{0.9\textwidth}{!}{%
\begin{tabular}{l|cccc|cccc}
\toprule
\multirow{2}{*}{\textbf{Method}} 
& \multicolumn{4}{c|}{\textbf{BCVL}} 
& \multicolumn{4}{c}{\textbf{IMEB}} \\
\cmidrule(lr){2-5} \cmidrule(lr){6-9}
& \#Tok(k) $\downarrow$ & \#Tool $\downarrow$ & Acc $\uparrow$ & CAS $\uparrow$ 
& \#Tok(k) $\downarrow$ & \#Tool $\downarrow$ & Acc $\uparrow$ & CAS $\uparrow$ \\
\midrule
MMSearch-R1      &   2.6 &  1.71 & 19.1 & 0.520
                 &   3.4 &  1.90 & 3.3 & 0.013 \\
DeepEyes-V2      &  24.2 &  4.30 & 24.8 & 0.182 
                 &  16.7 &  4.71 & 18.0 & 0.119 \\
WebWatcher       &  27.4 &  4.87 & 27.0 & 0.191 
                 &  22.8 &  7.82 & 15.3 & 0.059 \\
VDR              & 200.8 & 11.72 & 53.7 & 0.128 
                 & 303.4 & 12.34 & 21.2 & 0.014 \\
\midrule
\rowcolor{ours2}
\textbf{HyperEyes-30B} 
&  8.8 & 2.61 & \textbf{57.9} & \textbf{2.232}
& 16.7 & 3.13 & \textbf{46.7} & \textbf{0.910} \\
\bottomrule
\end{tabular}%
}
\arrayrulecolor{black}
\end{table*}

\subsection{Ablation Study}

\paragraph{Rigorous quality filtering dominates raw data volume.} 
We evaluate our data curation pipeline on the Qwen3-VL-30B backbone (Table~\ref{tab:abla_data}). The baseline utilizes 121,000 trajectories synthesized via progressive rejection sampling. Applying our strict trajectory-level filtering reduces the data volume to a quarter but improves average accuracy by 7.2 absolute points. These substantial gains on complex reasoning benchmarks prove that rigorous quality filtering dominates raw data volume for training efficient multimodal agents.

\paragraph{Adaptive efficiency references suppress redundancy and boost accuracy.} 
Table~\ref{tab:abla_results} ablates the reinforcement learning reward. Relying solely on outcome rewards inflates tool calls without improving baseline accuracy. Introducing a static tool-call reference resolves this inefficiency, lifting accuracy and restoring concise tool usage. Furthermore, the adaptive TRACE formulation creates a co-evolving curriculum that yields an additional 1.6 point accuracy increase and further minimizes tool calls, confirming the superiority of dynamic efficiency constraints.

\paragraph{Effective distillation strictly requires an efficiency-aligned teacher.} 
Finally, we evaluate on-policy distillation and teacher model selection (Table~\ref{tab:abla_results}). Augmenting TRACE with distillation from our aligned HyperEyes-235B model provides dense token-level supervision on failed rollouts, contributing an additional 1.3 point average accuracy gain without increasing the tool-call budget. Crucially, replacing this teacher with a vanilla Qwen3-VL-235B causes a severe accuracy drop. Moreover, skipping the initial fine-tuning phase entirely collapses performance, demonstrating that effective distillation strictly requires a parallel-amenable cold start and an efficiency-aligned teacher.

\begin{table*}[t]
\centering
\caption{Ablation on the training data curation pipeline (Accuracy \% / Avg. Tool-Call Turns).}
\label{tab:abla_data}
\renewcommand{\arraystretch}{1.15}
\resizebox{0.9\textwidth}{!}{%
\begin{tabular}{l|c|cccccc|c}
\toprule
\textbf{Data Variant} & \textbf{\#Samples} & \textbf{MMSearch} & \textbf{FVQA} & \textbf{LiveVQA}
& \textbf{BCVL} & \textbf{MMSearch+} & \textbf{IMEB} & \textbf{Avg.} \\
\midrule

\rowcolor{groupgray}
$\mathcal{D}_{\text{Base}}$
& 121k
& 77.9 / \textcolor{toolblue}{1.5} & 76.5 / \textcolor{toolblue}{1.6} & 74.4 / \textcolor{toolblue}{1.7}
& 38.8 / \textcolor{toolblue}{2.8} & 14.0 / \textcolor{toolblue}{2.9} & 28.0 / \textcolor{toolblue}{2.4}
& 51.6 / \textcolor{toolblue}{2.2} \\

\rowcolor{ourrow}
$\mathcal{D}_{\text{Filtering}}$ 
& 30k
& \textbf{82.0} / \textcolor{toolblue}{1.8} & \textbf{76.1} / \textcolor{toolblue}{2.0} & \textbf{80.3} / \textcolor{toolblue}{1.9}
& \textbf{47.6} / \textcolor{toolblue}{3.9} & \textbf{25.0} / \textcolor{toolblue}{3.7} & \textbf{42.0} / \textcolor{toolblue}{3.8}
& \textbf{58.8} / \textcolor{toolblue}{2.9} \\

\bottomrule
\end{tabular}}
\end{table*}
\begin{table*}[t]
\centering
\caption{Ablation on the Dual-Grained Efficiency-Aware RL (Accuracy \% / Avg. Tool-Call Turns). Direct RL denotes applying TRACE + OPD directly on the vanilla Qwen3-VL-30B-Instruct backbone without the SFT cold-start. For the OPD teacher: $^{\dagger}$ uses the off-the-shelf Qwen3-VL-235B-Instruct, while $^{\ast}$ (our default) uses the RL-trained HyperEyes-235B (RL).}
\label{tab:abla_results}
\renewcommand{\arraystretch}{1.15}
\resizebox{0.9\textwidth}{!}{%
\begin{tabular}{l|cccccc|c}
\toprule
\textbf{Model Variant} & \textbf{MMSearch} & \textbf{FVQA} & \textbf{LiveVQA}
& \textbf{BCVL} & \textbf{MMSearch+} & \textbf{IMEB} & \textbf{Avg.} \\
\midrule

\rowcolor{groupgray}
\textbf{Qwen3-VL-30B}
& 54.1 / \textcolor{toolblue}{1.7} & 58.0 / \textcolor{toolblue}{2.0} & 49.8 / \textcolor{toolblue}{1.9} & 29.0 / \textcolor{toolblue}{4.4} & 9.7 / \textcolor{toolblue}{2.8}  & 17.7 / \textcolor{toolblue}{4.3}  & 36.4 / \textcolor{toolblue}{2.7} \\

\quad {+} TRACE + OPD$^{\ast}$
& 64.8 / \textcolor{toolblue}{2.0} & 63.3 / \textcolor{toolblue}{1.7} & 65.6 / \textcolor{toolblue}{1.7} & 39.3 / \textcolor{toolblue}{2.7} & 15.3 / \textcolor{toolblue}{2.7} & 20.0 / \textcolor{toolblue}{2.4} & 44.7 / \textcolor{toolblue}{2.2} \\
\midrule
\rowcolor{groupgray}
\textbf{Qwen3-VL-30B + SFT}
& 82.0 / \textcolor{toolblue}{1.8} & 76.1 / \textcolor{toolblue}{2.0} & 80.3 / \textcolor{toolblue}{1.9} & 47.6 / \textcolor{toolblue}{3.9} & 25.0 / \textcolor{toolblue}{3.7} & 42.0 / \textcolor{toolblue}{3.8} & 58.8 / \textcolor{toolblue}{2.9} \\
\midrule

\quad {+} Outcome Reward
& 77.9 / \textcolor{toolblue}{6.8} & 73.4 / \textcolor{toolblue}{7.1} & 79.1 / \textcolor{toolblue}{6.0} & 52.4 / \textcolor{toolblue}{7.6} & 29.9 / \textcolor{toolblue}{7.0} & 37.3 / \textcolor{toolblue}{6.7} & 58.3 / \textcolor{toolblue}{6.9} \\

\quad {+} TRACE (w/o update)
& 84.4 / \textcolor{toolblue}{1.8} & 78.2 / \textcolor{toolblue}{1.8} & 81.6 / \textcolor{toolblue}{1.9} & 53.1 / \textcolor{toolblue}{2.9} & 27.8 / \textcolor{toolblue}{2.9} & 41.3 / \textcolor{toolblue}{3.7} & 61.1 / \textcolor{toolblue}{2.5} \\

\rowcolor{ourrow}
\quad {+} TRACE
& 84.4 / \textcolor{toolblue}{1.6} & 79.8 / \textcolor{toolblue}{1.7} & 82.4 / \textcolor{toolblue}{1.7} & 55.2 / \textcolor{toolblue}{2.5} & 29.2 / \textcolor{toolblue}{2.4} & 45.0 / \textcolor{toolblue}{3.3} & 62.7 / \textcolor{toolblue}{2.2} \\

\quad {+} TRACE + OPD$^{\dagger}$
& 68.9 / \textcolor{toolblue}{2.3} & 66.0 / \textcolor{toolblue}{1.8} & 64.4 / \textcolor{toolblue}{1.9} & 34.5 / \textcolor{toolblue}{4.1} & 13.9 / \textcolor{toolblue}{3.7} & 30.3 / \textcolor{toolblue}{2.9} & 46.3 / \textcolor{toolblue}{2.8} \\

\rowcolor{ourrow}
\quad {+} TRACE + OPD$^{\ast}$
& 86.9 / \textcolor{toolblue}{1.6} & 79.3 / \textcolor{toolblue}{1.7} & 81.6 / \textcolor{toolblue}{1.7} & 57.9 / \textcolor{toolblue}{2.6} & 31.5 / \textcolor{toolblue}{2.3} & 46.7 / \textcolor{toolblue}{3.1} & 64.0 / \textcolor{toolblue}{2.2} \\

\bottomrule
\end{tabular}}
\end{table*}

\section{Conclusion}

We present {HyperEyes}, a {parallel} multimodal search
agent that follows {search wider, not longer}: it fuses visual grounding and retrieval into a single atomic action, dispatching grounded queries concurrently within a round. Training proceeds in two stages. First, a {Parallel-Amenable Data Synthesis Pipeline} produces visual multi-entity and textual multi-constraint queries, from which {Progressive Rejection Sampling} and quality filtering distill $30$K efficient cold-start trajectories. Second, our central contribution, a {Dual-Grained Efficiency-Aware RL} framework, couples a macro-level reward ({TRACE}, with a monotonically tightened cost reference) with a micro-level signal ({OPD}, dense token-level supervision on failed rollouts). We further release {IMEB}, a $300$-instance benchmark scoring capability and efficiency. Across six benchmarks, HyperEyes-30B Pareto-dominates the strongest comparable-scale open-source agent by $\boldsymbol{+9.9}$ accuracy with $\boldsymbol{5.3{\times}}$ fewer tool-call rounds, and HyperEyes-235B closes to within $1.1$ accuracy points of Gemini-3.1-Pro, indicating that accuracy and efficiency are {complementary} rather than conflicting objectives in agentic multimodal search.

\clearpage
\bibliographystyle{neurips_2023}
\bibliography{reference}

@inproceedings{brown2020language,
  author       = {Tom B. Brown and
                  Benjamin Mann and
                  Nick Ryder and
                  Melanie Subbiah and
                  Jared Kaplan and
                  Prafulla Dhariwal and
                  Arvind Neelakantan and
                  Pranav Shyam and
                  Girish Sastry and
                  Amanda Askell and
                  Sandhini Agarwal and
                  Ariel Herbert{-}Voss and
                  Gretchen Krueger and
                  Tom Henighan and
                  Rewon Child and
                  Aditya Ramesh and
                  Daniel M. Ziegler and
                  Jeffrey Wu and
                  Clemens Winter and
                  Christopher Hesse and
                  Mark Chen and
                  Eric Sigler and
                  Mateusz Litwin and
                  Scott Gray and
                  Benjamin Chess and
                  Jack Clark and
                  Christopher Berner and
                  Sam McCandlish and
                  Alec Radford and
                  Ilya Sutskever and
                  Dario Amodei},
  title        = {Language Models are Few-Shot Learners},
  booktitle    = {Advances in Neural Information Processing Systems 33: Annual Conference
                  on Neural Information Processing Systems 2020, NeurIPS 2020, December
                  6-12, 2020, virtual},
  year         = {2020},
  url          = {https://proceedings.neurips.cc/paper/2020/hash/1457c0d6bfcb4967418bfb8ac142f64a-Abstract.html},

}

@article{gemini2024,
  author       = {Machel Reid and
                  Nikolay Savinov and
                  Denis Teplyashin and
                  Dmitry Lepikhin and
                  Timothy P. Lillicrap and
                  Jean{-}Baptiste Alayrac and
                  Radu Soricut and
                  Angeliki Lazaridou and
                  Orhan Firat and
                  Julian Schrittwieser and
                  Ioannis Antonoglou and
                  Rohan Anil and
                  Sebastian Borgeaud and
                  Andrew M. Dai and
                  Katie Millican and
                  Ethan Dyer and
                  Mia Glaese and
                  Thibault Sottiaux and
                  Benjamin Lee and
                  Fabio Viola and
                  Malcolm Reynolds and
                  Yuanzhong Xu and
                  James Molloy and
                  Jilin Chen and
                  Michael Isard and
                  Paul Barham and
                  Tom Hennigan and
                  Ross McIlroy and
                  Melvin Johnson and
                  Johan Schalkwyk and
                  Eli Collins and
                  Eliza Rutherford and
                  Erica Moreira and
                  Kareem Ayoub and
                  Megha Goel and
                  Clemens Meyer and
                  Gregory Thornton and
                  Zhen Yang and
                  Henryk Michalewski and
                  Zaheer Abbas and
                  Nathan Schucher and
                  Ankesh Anand and
                  Richard Ives and
                  James Keeling and
                  Karel Lenc and
                  Salem Haykal and
                  Siamak Shakeri and
                  Pranav Shyam and
                  Aakanksha Chowdhery and
                  Roman Ring and
                  Stephen Spencer and
                  Eren Sezener and
                  et al.},
  title        = {Gemini 1.5: Unlocking multimodal understanding across millions of
                  tokens of context},
  journal      = {CoRR},
  volume       = {abs/2403.05530},
  year         = {2024},
  url          = {https://doi.org/10.48550/arXiv.2403.05530},
  doi          = {10.48550/ARXIV.2403.05530},
  eprinttype    = {arXiv},
  eprint       = {2403.05530},
  bibsource    = {dblp computer science bibliography, https://dblp.org}
}

@article{hurst2024gpt,
  title={Gpt-4o system card},
  author={Hurst, Aaron and Lerer, Adam and Goucher, Adam P and Perelman, Adam and Ramesh, Aditya and Clark, Aidan and Ostrow, AJ and Welihinda, Akila and Hayes, Alan and Radford, Alec and others},
  journal={arXiv preprint arXiv:2410.21276},
  year={2024}
}

@article{ouyang2022training,
  title={Training language models to follow instructions with human feedback},
  author={Ouyang, Long and Wu, Jeffrey and Jiang, Xu and Almeida, Diogo and Wainwright, Carroll and Mishkin, Pamela and Zhang, Chong and Agarwal, Sandhini and Slama, Katarina and Ray, Alex and others},
  journal={Advances in neural information processing systems},
  volume={35},
  pages={27730--27744},
  year={2022}
}

@article{anil2023palm,
  title={Palm 2 technical report},
  author={Anil, Rohan and Dai, Andrew M and Firat, Orhan and Johnson, Melvin and Lepikhin, Dmitry and Passos, Alexandre and Shakeri, Siamak and Taropa, Emanuel and Bailey, Paige and Chen, Zhifeng and others},
  journal={arXiv preprint arXiv:2305.10403},
  url          = {https://doi.org/10.48550/arXiv.2305.10403},
  year={2023}
}

@article{wu2025mmsearch,
  title={Mmsearch-r1: Incentivizing lmms to search},
  author={Wu, Jinming and Deng, Zihao and Li, Wei and Liu, Yiding and You, Bo and Li, Bo and Ma, Zejun and Liu, Ziwei},
  journal={arXiv preprint arXiv:2506.20670},
  year={2025}
}

@article{tao2025mmsearch,
  title={Mmsearch-plus: Benchmarking provenance-aware search for multimodal browsing agents},
  author={Tao, Xijia and Teng, Yihua and Su, Xinxing and Fu, Xinyu and Wu, Jihao and Tao, Chaofan and Liu, Ziru and Bai, Haoli and Liu, Rui and Kong, Lingpeng},
  journal={arXiv preprint arXiv:2508.21475},
  year={2025}
}

@article{geng2025webwatcher,
  title={Webwatcher: Breaking new frontier of vision-language deep research agent},
  author={Geng, Xinyu and Xia, Peng and Zhang, Zhen and Wang, Xinyu and Wang, Qiuchen and Ding, Ruixue and Wang, Chenxi and Wu, Jialong and Zhao, Yida and Li, Kuan and others},
  journal={arXiv preprint arXiv:2508.05748},
  year={2025}
}

@article{jiang2024mmsearch,
  title={Mmsearch: Benchmarking the potential of large models as multi-modal search engines},
  author={Jiang, Dongzhi and Zhang, Renrui and Guo, Ziyu and Wu, Yanmin and Lei, Jiayi and Qiu, Pengshuo and Lu, Pan and Chen, Zehui and Fu, Chaoyou and Song, Guanglu and others},
  journal={arXiv preprint arXiv:2409.12959},
  year={2024}
}

@article{fu2025seeking,
  title={Seeking and updating with live visual knowledge},
  author={Fu, Mingyang and Peng, Yuyang and Chen, Dongping and Zhou, Zetong and Liu, Benlin and Wan, Yao and Zhao, Zhou and Yu, Philip S and Krishna, Ranjay},
  journal={arXiv preprint arXiv:2504.05288},
  year={2025}
}

@inproceedings{
jin2025searchr,
title={Search-R1: Training {LLM}s to Reason and Leverage Search Engines with Reinforcement Learning},
author={Bowen Jin and Hansi Zeng and Zhenrui Yue and Jinsung Yoon and Sercan O Arik and Dong Wang and Hamed Zamani and Jiawei Han},
booktitle={Second Conference on Language Modeling},
year={2025},
url={https://openreview.net/forum?id=Rwhi91ideu}
}

@inproceedings{
yao2023react,
title={ReAct: Synergizing Reasoning and Acting in Language Models},
author={Shunyu Yao and Jeffrey Zhao and Dian Yu and Nan Du and Izhak Shafran and Karthik R Narasimhan and Yuan Cao},
booktitle={The Eleventh International Conference on Learning Representations },
year={2023},
url={https://openreview.net/forum?id=WE_vluYUL-X}
}

@article{huang2026vision,
  title={Vision-deepresearch: Incentivizing deepresearch capability in multimodal large language models},
  author={Huang, Wenxuan and Zeng, Yu and Wang, Qiuchen and Fang, Zhen and Cao, Shaosheng and Chu, Zheng and Yin, Qingyu and Chen, Shuang and Yin, Zhenfei and Chen, Lin and others},
  journal={arXiv preprint arXiv:2601.22060},
  year={2026}
}

@article{chu2026redsearcher,
  title={Redsearcher: A scalable and cost-efficient framework for long-horizon search agents},
  author={Chu, Zheng and Wang, Xiao and Hong, Jack and Fan, Huiming and Huang, Yuqi and Yang, Yue and Xu, Guohai and Zhao, Chenxiao and Xiang, Cheng and Hu, Shengchao and others},
  journal={arXiv preprint arXiv:2602.14234},
  year={2026}
}

@article{team2026kimi,
  title={Kimi K2. 5: Visual Agentic Intelligence},
  author={Team, Kimi and Bai, Tongtong and Bai, Yifan and Bao, Yiping and Cai, SH and Cao, Yuan and Charles, Y and Che, HS and Chen, Cheng and Chen, Guanduo and others},
  journal={arXiv preprint arXiv:2602.02276},
  year={2026}
}

@article{bai2025qwen3,
  title={Qwen3-vl technical report},
  author={Bai, Shuai and Cai, Yuxuan and Chen, Ruizhe and Chen, Keqin and Chen, Xionghui and Cheng, Zesen and Deng, Lianghao and Ding, Wei and Gao, Chang and Ge, Chunjiang and others},
  journal={arXiv preprint arXiv:2511.21631},
  year={2025}
}

@article{zhao2025parallelsearch,
  title={Parallelsearch: Train your llms to decompose query and search sub-queries in parallel with reinforcement learning},
  author={Zhao, Shu and Yu, Tan and Xu, Anbang and Singh, Japinder and Shukla, Aaditya and Akkiraju, Rama},
  journal={arXiv preprint arXiv:2508.09303},
  year={2025}
}

@article{lin2026w,
  title={W\&D: Scaling Parallel Tool Calling for Efficient Deep Research Agents},
  author={Lin, Xiaoqiang and Liew, Jun Hao and Savarese, Silvio and Li, Junnan},
  journal={arXiv preprint arXiv:2602.07359},
  year={2026}
}

@inproceedings{
ko2026hybrid,
title={Hybrid Deep Searcher: Scalable Parallel and Sequential Search Reasoning},
author={Dayoon Ko and Jihyuk Kim and Haeju Park and Sohyeon Kim and Dahyun Lee and Yongrae Jo and Gunhee Kim and Moontae Lee and Kyungjae Lee},
booktitle={The Fourteenth International Conference on Learning Representations},
year={2026},
url={https://openreview.net/forum?id=rXpTZyucal}
}

@inproceedings{
hong2026deepeyesv,
title={DeepEyesV2: Toward Agentic Multimodal Model},
author={Jack Hong and Chenxiao Zhao and ChengLIn Zhu and Weiheng Lu and Guohai Xu and XingYu},
booktitle={The Fourteenth International Conference on Learning Representations},
year={2026},
url={https://openreview.net/forum?id=yDKawwfJ5O}
}

@article{lu2025deepdive,
  title={Deepdive: Advancing deep search agents with knowledge graphs and multi-turn rl},
  author={Lu, Rui and Hou, Zhenyu and Wang, Zihan and Zhang, Hanchen and Liu, Xiao and Li, Yujiang and Feng, Shi and Tang, Jie and Dong, Yuxiao},
  journal={arXiv preprint arXiv:2509.10446},
  year={2025}
}

@article{marsili2025same,
  title={Same or Not? Enhancing Visual Perception in Vision-Language Models},
  author={Marsili, Damiano and Mehta, Aditya and Lin, Ryan Y and Gkioxari, Georgia},
  journal={arXiv preprint arXiv:2512.23592},
  year={2025}
}

@article{lewis2020retrieval,
  title={Retrieval-augmented generation for knowledge-intensive nlp tasks},
  author={Lewis, Patrick and Perez, Ethan and Piktus, Aleksandra and Petroni, Fabio and Karpukhin, Vladimir and Goyal, Naman and K{\"u}ttler, Heinrich and Lewis, Mike and Yih, Wen-tau and Rockt{\"a}schel, Tim and others},
  journal={Advances in neural information processing systems},
  volume={33},
  pages={9459--9474},
  year={2020}
}

@article{vrandevcic2014wikidata,
  title={Wikidata: a free collaborative knowledgebase},
  author={Vrande{\v{c}}i{\'c}, Denny and Kr{\"o}tzsch, Markus},
  journal={Communications of the ACM},
  volume={57},
  number={10},
  pages={78--85},
  year={2014},
  publisher={ACM New York, NY, USA}
}

@techreport{WahCUB_200_2011,
	Author = {Wah, C. and Branson, S. and Welinder, P. and Perona, P. and Belongie, S.},
	Institution = {California Institute of Technology},
	Number = {CNS-TR-2011-001},
	Title = {{The Caltech-UCSD Birds-200-2011 Dataset}},
	Year = {2011}
}

@InProceedings{Nilsback08,
  author       = "Nilsback, M.-E. and Zisserman, A.",
  title        = "Automated Flower Classification over a Large Number of Classes",
  booktitle    = "Proceedings of the Indian Conference on Computer Vision, Graphics and Image Processing",
  year         = "2008",
  month        = "Dec"
}

@inproceedings{KrauseStarkDengFei-Fei_3DRR2013,
  title = {3D Object Representations for Fine-Grained Categorization},
  booktitle = {IEEE Workshop on 3D Representation and Recognition},
  author = {Krause, Jonathan and Stark, Michael and Deng, Jia and Fei-Fei, Li},
  year = {2013}
}

@article{maji2013fine,
  title={Fine-grained visual classification of aircraft},
  author={Maji, Subhransu and Kannala, Juho and Rahtu, Esa and Blaschko, Matthew and Vedaldi, Andrea},
  journal={arXiv preprint arXiv:1306.5151},
  year={2013}
}

@InProceedings{parkhi12a,
  author       = "Parkhi, O. M. and Vedaldi, A. and Zisserman, A. and Jawahar, C. V.",
  title        = "Cats and Dogs",
  booktitle    = "IEEE Conference on Computer Vision and Pattern Recognition",
  year         = "2012",
}

@article{zhao2024swift,
  title   = {{SWIFT}: A Scalable lightWeight Infrastructure for Fine-Tuning},
  author  = {Zhao, Yuze and Huang, Jintao and Hu, Jinghan and Wang, Xingjun
             and Mao, Yunlin and Zhang, Daoze and Zhang, Hong and Jiang, Zeyinzi
             and Wu, Zhikai and Ai, Baole and Wang, Ang and Zhou, Wenmeng
             and Chen, Yingda},
  journal = {arXiv preprint arXiv:2408.05517},
  year    = {2024},
  url     = {https://arxiv.org/abs/2408.05517}
}

@misc{relax2026,
  title        = {{Relax}: An Asynchronous Reinforcement Learning Framework
                  for Large-Scale Agentic Models},
  author       = {{Relax Contributors}},
  year         = {2026},
  howpublished = {\url{https://github.com/redai-infra/Relax}},
  note         = {Open-source software}
}

@inproceedings{zheng2024sglang,
  title     = {{SGLang}: Efficient Execution of Structured Language Model Programs},
  author    = {Zheng, Lianmin and Yin, Liangsheng and Xie, Zhiqiang and Sun, Chuyue
               and Huang, Jeff and Yu, Cody Hao and Cao, Shiyi and Kozyrakis, Christos
               and Stoica, Ion and Gonzalez, Joseph E. and Barrett, Clark and Sheng, Ying},
  booktitle = {NeurIPS},
  year      = {2024}
}

@article{shoeybi2019megatron,
  title   = {{Megatron-LM}: Training Multi-Billion Parameter Language Models
             Using Model Parallelism},
  author  = {Shoeybi, Mohammad and Patwary, Mostofa and Puri, Raul and LeGresley, Patrick
             and Casper, Jared and Catanzaro, Bryan},
  journal = {arXiv preprint arXiv:1909.08053},
  year    = {2019}
}

@article{shao2024deepseekmathpushinglimitsmathematical,
 author = {Shao, Zhihong and Wang, Peiyi and Zhu, Qihao and Xu, Runxin and Song, Junxiao and Bi, Xiao and Zhang, Haowei and Zhang, Mingchuan and Li, YK and Wu, Y and others},
 journal = {ArXiv preprint},
 title = {Deepseekmath: Pushing the limits of mathematical reasoning in open language models},
 url = {https://arxiv.org/abs/2402.03300},
 volume = {abs/2402.03300},
 year = {2024}
}

@inproceedings{
gu2024minillm,
title={Mini{LLM}: Knowledge Distillation of Large Language Models},
author={Yuxian Gu and Li Dong and Furu Wei and Minlie Huang},
booktitle={The Twelfth International Conference on Learning Representations},
year={2024},
url={https://openreview.net/forum?id=5h0qf7IBZZ}
}

@inproceedings{press-etal-2023-measuring,
    title = "Measuring and Narrowing the Compositionality Gap in Language Models",
    author = "Press, Ofir  and
      Zhang, Muru  and
      Min, Sewon  and
      Schmidt, Ludwig  and
      Smith, Noah  and
      Lewis, Mike",
    editor = "Bouamor, Houda  and
      Pino, Juan  and
      Bali, Kalika",
    booktitle = "Findings of the Association for Computational Linguistics: EMNLP 2023",
    month = dec,
    year = "2023",
    address = "Singapore",
    publisher = "Association for Computational Linguistics",
    url = "https://aclanthology.org/2023.findings-emnlp.378/",
    doi = "10.18653/v1/2023.findings-emnlp.378",
    pages = "5687--5711"
}

@article{team2025mirothinker,
  title={Mirothinker: Pushing the performance boundaries of open-source research agents via model, context, and interactive scaling},
  author={Team, MiroMind and Bai, Song and Bing, Lidong and Chen, Carson and Chen, Guanzheng and Chen, Yuntao and Chen, Zhe and Chen, Ziyi and Dai, Jifeng and Dong, Xuan and others},
  journal={arXiv preprint arXiv:2511.11793},
  year={2025}
}

@article{zhang2025survey,
  title={A survey on test-time scaling in large language models: What, how, where, and how well?},
  author={Zhang, Qiyuan and Lyu, Fuyuan and Sun, Zexu and Wang, Lei and Zhang, Weixu and Hua, Wenyue and Wu, Haolun and Guo, Zhihan and Wang, Yufei and Muennighoff, Niklas and others},
  journal={arXiv preprint arXiv:2503.24235},
  year={2025}
}

@article{narayan2025deepmmsearch,
  title={Deepmmsearch-r1: Empowering multimodal llms in multimodal web search},
  author={Narayan, Kartik and Xu, Yang and Cao, Tian and Nerella, Kavya and Patel, Vishal M and Shiee, Navid and Grasch, Peter and Jia, Chao and Yang, Yinfei and Gan, Zhe},
  journal={arXiv preprint arXiv:2510.12801},
  year={2025}
}

@article{zhang2025skywork,
  title={Skywork-R1V4: Toward Agentic Multimodal Intelligence through Interleaved Thinking with Images and DeepResearch},
  author={Zhang, Yifan and Hu, Liang and Sun, Haofeng and Wang, Peiyu and Wei, Yichen and Yin, Shukang and Pei, Jiangbo and Shen, Wei and Xia, Peng and Peng, Yi and others},
  journal={arXiv preprint arXiv:2512.02395},
  year={2025}
}

@article{xia2025open,
  title={Open data synthesis for deep research},
  author={Xia, Ziyi and Luo, Kun and Qian, Hongjin and Liu, Zheng},
  journal={arXiv preprint arXiv:2509.00375},
  year={2025}
}

@inproceedings{
vendrow2024inquire,
title={{INQUIRE}: A Natural World Text-to-Image Retrieval Benchmark},
author={Edward Vendrow and Omiros Pantazis and Alexander Shepard and Gabriel Brostow and Kate E. Jones and Oisin Mac Aodha and Sara Beery and Grant Van Horn},
booktitle={The Thirty-eight Conference on Neural Information Processing Systems Datasets and Benchmarks Track},
year={2024},
url={https://openreview.net/forum?id=jbrMS0DNaD}
}

@inproceedings{49052,title	= {Google Landmarks Dataset v2 - A Large-Scale Benchmark for Instance-Level Recognition and Retrieval},author	= {Tobias Weyand and André Araujo and Bingyi Cao and Jack Sim},year	= {2020},URL	= {https://arxiv.org/abs/2004.01804},booktitle	= {CVPR}}

@misc{gemini2026pro,
  title        = {{Gemini 3.1 Pro: A smarter model for your most complex tasks}},
  author       = {Google DeepMind},
  year         = {2026},
  month        = feb,
  url          = {https://blog.google/innovation-and-ai/models-and-research/gemini-models/gemini-3-1-pro/},
  note         = {Accessed: 2026-05-07}
}

@misc{anthropic2026opus,
  title        = {{Introducing Claude Opus 4.6}},
  author       = {Anthropic},
  year         = {2026},
  month        = feb,
  url          = {https://www.anthropic.com/news/claude-opus-4-6},
  note         = {Accessed: 2026-05-07}
}

\clearpage

\appendix
\section{Limitations}
\label{appendix:limitations}
While HyperEyes establishes a robust baseline for efficient multimodal search, we identify several limitations. First, On-Policy Distillation requires a stronger same-family teacher, which inherently bounds the student's reasoning capabilities and prevents direct application at the frontier scale. Second, our parallel framework focuses exclusively on static image and text environments, lacking the spatial-temporal grounding mechanisms necessary for dynamic modalities like video or audio. Finally, a residual performance gap persists compared to leading closed-source frontier models (e.g., Gemini-3.1-Pro), highlighting the need for larger-scale reinforcement learning and more diverse multimodal training distributions in future research.

\section{Broader Impacts}
\label{app:broader_impacts}

\paragraph{Efficiency and accessibility.} HyperEyes reduces the per-query tool cost of multimodal search agents by roughly $5\times$ at comparable or better accuracy. This translates directly into lower energy consumption and shorter end-to-end latency for downstream multimodal assistants, making real-time grounded multimodal QA accessible to users on resource-constrained devices and to researchers without hyperscale compute.

\paragraph{Open and reproducible research.} By releasing the IMEB benchmark, the Parallel-Amenable Data Synthesis pipeline, the full SFT and RL training recipes, and the HyperEyes-30B checkpoints, this work establishes a reproducible foundation for the community to study and improve efficiency-aware multimodal agents, lowering the entry barrier for academic and resource-constrained groups.

\paragraph{Enabling grounded multimodal applications.} The Unified Grounded Search action space and Dual-Grained Efficiency-Aware RL framework offer a general recipe for building reliable multimodal assistants in domains where verifiable, source-grounded answers are essential, such as education, scientific literature exploration, accessibility tools for visually impaired users, and consumer-facing visual question answering.

\section{Dataset Curation Details}
\label{appendix:data-curation}

\subsection{Parallel-Amenable QA Synthesis}
\subsubsection{Details of Visual Multi-Entity Synthesis}

The visual multi-entity synthesis pipeline consists of three steps: source data selection, per-class knowledge base and QA pool construction, and mosaic-based multi-entity QA composition.

\paragraph{Source data.}
We adopt five fine-grained classification datasets as the base corpus, covering birds (CUB-200-2011~\cite{WahCUB_200_2011}), flowers (Oxford Flowers-102~\cite{Nilsback08}), cars (Stanford Cars~\cite{KrauseStarkDengFei-Fei_3DRR2013}), aircraft (FGVC-Aircraft~\cite{maji2013fine}), and pets (Oxford-IIIT Pets~\cite{parkhi12a}). These datasets provide only images and class names, lacking the fine-grained world knowledge needed to construct retrieval-oriented QA, which motivates the introduction of external information described next.

\paragraph{Per-class knowledge base and QA pool.}
For each class name, we invoke Gemini-3.0-Flash to perform web search, page crawling, and information aggregation, covering dimensions such as visual characteristics, taxonomic identifiers, historical background, and ecological or behavioral traits, and consolidate the results into a structured knowledge entry per class. On top of this, we randomly sample 2 to 8 entries from each class as reference evidence and prompt Gemini-3.0-Flash to synthesize the corresponding QA pool, requiring questions to use unambiguous referents, and answers to be concise phrases, entity names, or numeric values, while avoiding open-ended formulations and meta-references to the source documents. Each class in the resulting knowledge base is therefore characterized jointly by its original images, structured knowledge entries, and the associated QA pool, providing reusable fine-grained question-answer material for the subsequent mosaic stage.

\paragraph{Mosaic-based multi-entity QA composition.}
To produce the final training samples, we randomly draw images from 2 to 8 classes within the same source dataset, allowing repetition, and assemble them into a composite image via regular-grid mosaicking with layouts such as $2{\times}1$, $1{\times}4$, and $2{\times}4$. We then sample several knowledge snippets from each selected class's QA pool, and feed them, together with the spatial position of every target entity in the composite (e.g., ``top-left'' or ``second in the first row''), to Gemini-3.0-Flash, which fuses the spatial references with the knowledge content to generate the final multi-entity QA. This construction ensures that answering each question requires concurrently localizing and retrieving knowledge about multiple target entities, since omitting any single entity yields an incomplete answer, thereby imposing a data-level constraint that drives the model to learn parallel retrieval behavior. After the entire pipeline, the visual multi-entity corpus contains $20{,}000$ QA pairs in total.

\subsubsection{Details of Textual Multi-Constraint Synthesis} 

The pipeline of Textual Multi-Constraint Synthesis consists of four steps: pivot discovery, attribute filtering, constraint-chain construction, and natural-language question generation.

\paragraph{Pivot discovery and candidate set construction.}
Starting from a randomly sampled seed entity, we perform a 2 to 3 hop random walk over Wikidata~\cite{vrandevcic2014wikidata} to reach a pivot entity $A$, and record the full traversal path so that it can be directly reused as the reasoning chain of a downstream multi-hop QA. We then collect the first-order neighbors $\mathcal{N}(A)=\{B_i\}$ of $A$ as the candidate answer set, from which the attribute set used for subsequent predicate sampling is extracted. Following the setting in Sec.~\ref{sec:met_Data}, we retain only pivots with $|\mathcal{N}(A)|\in[4,12]$, so that the candidate pool is rich enough to support compositional filtering yet small enough to keep the eventual answer set tractable.

\paragraph{Attribute whitelist and blacklist.}
To suppress the low-information and high-bias constraints introduced by naive sampling over Wikidata, we maintain a whitelist of highly discriminative attributes covering categories such as occupation, education, achievements, film production, music, organizations, and events. The $m\!\geq\!2$ predicates required by Sec.\ref{subsubsec:qa} are sampled exclusively from this whitelist. In parallel, we explicitly forbid a set of high-bias attributes including geography, country, religion, and gender, in order to prevent questions that can be shortcut-solved via surface-level geographic or demographic correlations. At the value level, a constraint value is further discarded if it satisfies any of the following criteria: (i) it belongs to a manually curated set of overly abstract concepts such as ``human'' or ``organization''; (ii) its Wikidata out-degree exceeds 300, indicating a hub entity referenced by too many neighbors and thus offering little discriminative signal; or (iii) its own type falls into bias-prone categories such as country, city, political party, or religion.

\paragraph{Greedy constraint-chain construction.}
On the resulting candidate set, we greedily append predicates to progressively shrink the pool toward a target size of $|\mathcal{B}^*|\in[1,8]$. At each step, the next predicate is required to (i) yield a non-empty intersection with the current candidate set, and (ii) maximize a domain-diversity bonus that favors predicates drawn from a different attribute family than those already selected, thereby preventing homogeneous constraints. The procedure terminates as soon as the candidate set falls within the target range. As an illustration, consider a candidate set of twelve actors $\{\text{Murphy}, \text{Bale}, \text{Hardy}, \text{DiCaprio}, \text{Watts}, \dots\}$: applying the predicate ``occupation = actor'' reduces the set to nine candidates, and a cross-domain predicate ``award received = BAFTA'' further narrows it to three, at which point construction terminates. Combined with the whitelist priority and value-level filtering, this procedure ensures that every retained constraint chain is compositionally non-trivial and robust to surface-level shortcuts.

\paragraph{Natural-language question generation.} We invoke Gemini-3.0-Flash to obfuscate the key attributes and values for each retained constraint chain $(A,\{(\text{attr}_k,v_k)\}_{k=1}^m,\mathcal{B}^*)$, transforming the structured constraints into fluent natural-language questions. The prompt explicitly instructs the model to paraphrase predicates rather than verbatim enumerate attribute keys and values, so that the surface form does not directly leak the underlying schema. For instance, rather than listing the raw attribute names and values, the model produces natural phrasings such as ``directed by Christopher Nolan, released after 2010, and with a runtime exceeding 140 minutes''.

\subsubsection{QA Filtering}
\label{app:qa-filter}

To ensure that the final QA data used for training is factually accurate and genuinely retrieval-dependent, we apply a unified two-stage filtering procedure to all QA pairs synthesized by the two pipelines. In the first stage, we employ Gemini-3.0-Flash as a judge to score each candidate QA along six dimensions: factual consistency, answer uniqueness, phrasing clarity, temporal stability, linguistic naturalness, and answer non-leakage; any sample that fails on any dimension is discarded. In the second stage, we further evaluate the remaining QA pairs with Qwen3-VL-235B under a tool-free pass@1 setting: any sample that can be answered correctly solely from the model's parametric knowledge is removed, so that every retained QA truly requires external retrieval to be solved. After this filtering procedure, the textual multi-constraint corpus retains $5{,}000$ QA pairs and the visual multi-entity corpus retains $20{,}000$ QA pairs.

\subsection{SFT data}
\label{appendix:sft-data}
\begin{algorithm}[htbp]
\caption{Progressive Rejection Sampling}
\label{alg:prs}
\begin{algorithmic}[1]
\Require QA pair $(q,a^*)$; Agent policy $\pi$; Ascending budgets list $B$; Number of rollouts $K$; Verifier $\mathrm{Judge}$
\Ensure Shortest successful trajectory $\tau^*$ or \texttt{REJECT}

\Statex \textcolor{blue}{\textit{Step 1: Progressive Search over Budgets}}
\For{budget $b \in B$}
    
    \Statex \hspace{\algorithmicindent} \textcolor{blue}{\textit{Step 2: Trajectory Sampling}}
    \State $\mathcal{T}_b \sim \pi(\cdot \mid q,b)^{K}$ 
    \Comment{\textcolor{green!50!black}{Sample $K$ trajectories for prompt $q$ under budget $b$}}
    
    \Statex \hspace{\algorithmicindent} \textcolor{blue}{\textit{Step 3: Verification}}
    \State $\mathcal{T}_b^{+} \leftarrow \{\tau \in \mathcal{T}_b : \mathrm{Judge}(\tau,a^*) = 1\}$ 
    \Comment{\textcolor{green!50!black}{Filter to keep only successful trajectories}}
    
    \Statex \hspace{\algorithmicindent} \textcolor{blue}{\textit{Step 4: Optimal Selection}}
    \If{$\mathcal{T}_b^{+} \neq \emptyset$} 
        \Comment{\textcolor{green!50!black}{If at least one guided trajectory succeeds}}
        \State \Return $\arg\min_{\tau\in\mathcal{T}_b^{+}}\mathrm{Turns}(\tau)$ 
        \Comment{\textcolor{green!50!black}{Return the shortest one to optimize efficiency}}
    \EndIf
\EndFor

\Statex \textcolor{blue}{\textit{Default case: All budgets exhausted}}
\State \Return \texttt{REJECT} 
\Comment{\textcolor{green!50!black}{No successful trajectory found across all budgets}}

\end{algorithmic}
\end{algorithm}

To construct a high-quality and efficiency-oriented supervised fine-tuning dataset, we aggregate a diverse set of query-answer pairs and process them through our progressive rejection sampling pipeline. Table~\ref{tab:data_composition} summarizes the initial pool of 271,000 QA pairs. This raw corpus originates from three main sources: public benchmarks covering multi-hop reasoning and fine-grained visual recognition (e.g., LiveVQA, InfoSeek, iNaturalist), our self-synthesized parallel-amenable corpora, and a specialized human-annotated dataset.

\paragraph{Difficulty filtering.}
Before generating trajectories, we filter out trivial queries to maximize the learning density of the dataset. Specifically, we deploy the Qwen3-VL-235B model to answer all 271,000 queries with full tool access under a pass@1 evaluation setting. If the model successfully resolves a query on its first attempt, we consider the sample too easy and lacking sufficient difficulty to teach advanced parallel search strategies. We strictly discard these solvable instances and retain only the challenging queries that demand rigorous multi-step or parallel planning.

\paragraph{Progressive rejection sampling.}
For the retained challenging queries, we employ Gemini-3.0-Flash as the policy model to sample tool-use trajectories. As detailed in Algorithm~\ref{alg:prs}, we define an ascending schedule of maximum turn budgets $B = \{2, 4, 8\}$. At each budget level $b \in B$, the model samples five candidate trajectories in a pass@5 configuration. If the policy finds at least one successful trajectory within a tight budget, the algorithm retains the shortest successful rollout and immediately terminates sampling for larger budgets. By explicitly prioritizing success under minimal turn constraints, this mechanism systematically suppresses verbose sequential interactions and elicits the unified grounded search behavior, wherein the agent processes multiple entities concurrently.

\paragraph{Quality filtering.}
Finally, we subject the surviving shortest trajectories to stringent quality constraints to ensure the agent learns robust, parallelized reasoning rather than parametric guessing or suboptimal heuristics. We eliminate trajectories exhibiting any of the following flaws. First, format violations and reasoning omissions: we discard trajectories containing invalid JSON structures or failing to strictly adhere to the ``think-before-act'' paradigm, which requires a ``<reason>'' block to precede any ``<tool\_call>'' or ``<answer>''. Second, unimodal and sequential shortcuts: we discard samples that can be resolved solely via image search, as they fail to incentivize multimodal synergy. Furthermore, we strictly filter out trajectories that degenerate into inefficient sequential querying by failing to trigger concurrent operations during text retrieval. Third, zero information gain: we reject cases where the agent executes repetitive searches yielding duplicate web snippets or off-topic results. Fourth, ungrounded answers: we exclude trajectories where the final conclusion hallucinates facts or leaps to a correct answer without sufficient supporting evidence present in the retrieved context. Through this rigorous cascade of difficulty filtering, budget-constrained sampling, and quality control, we distill the initial 271,000 QA pairs into 30,000 highly efficient and zero-redundancy trajectories. This curated dataset serves as a high-fidelity cold-start demonstration for HyperEyes.

\subsection{RL data}
\label{appendix:rl-data}

To construct the reinforcement learning dataset, we deploy the SFT-trained model to re-evaluate the queries that previously failed during the progressive rejection sampling phase. We strictly filter these queries by retaining only those where the policy fails on its first attempt (pass@1) but successfully resolves the task within five attempts (pass@5). Ultimately, we selected samples of moderate difficulty—specifically, 6,056 samples for the 30B variant and 9,337 samples for the 235B variant—providing sufficient exploration space while ensuring an initial positive reward signal. For each retained query, we extracted the first successful trajectory from the iteration results of pass@5 and recorded the number of tool call rounds ($t_c$) and the total number of tool calls ($t_s$). These metrics directly serve as initial efficiency reference values $\hat{t}_c^{(0)}$ and $\hat{t}_s^{(0)}$ for the TRACE reward formula.

\subsection{IMEB data}
\label{appendix:IMEB-data}

As outlined in Section~\ref{sec:met_IMEB}, the curation of the Image Multi-Entity Benchmark~(IMEB) involves a rigorous manual annotation and automated filtering pipeline designed to assess parallel multimodal search capabilities. First, five active PhD students manually source diverse multi-entity images from the web. The annotators deliberately vary the dataset across entity counts, domain categories, question types, and visual complexities~(Figure~\ref{fig:IMEB}). For each image, they author complex question-answer pairs that are impossible to deduce solely from the raw visual content.

Following the initial drafting, the dataset undergoes multiple rounds of double-blind cross-validation among the annotators. During this phase, reviewers independently attempt to solve the proposed queries using search tools. This rigorous peer-review process ensures three critical properties for every instance: (1)~it is unambiguously solvable given the correct retrieved context; (2)~it is free of informational ambiguity or conflicting ground truths; and (3)~it strictly requires parallel search across multiple entities to be answered efficiently. Queries that can be trivially resolved via single-hop sequential search are iteratively revised or discarded.

Finally, to ensure the absolute necessity of external knowledge, candidates pass through a Qwen3-VL-235B verifier that explicitly removes ``tool-unnecessary'' questions. Specifically, any instance that the model answers correctly without invoking search tools is discarded. This dual-layered verification guarantees that every retained instance genuinely demands multi-entity concurrent search, establishing a highly credible testbed for efficiency-aware agentic evaluations.

\section{Evaluation Details}
\label{sec:evaluation_details}

To ensure a fair and comprehensive evaluation, it is crucial to recognize the inherent inconsistencies in the deployment environments and toolchains employed by various open-source baselines. Table~\ref{tab:tool_inconsistency} summarizes the discrepancies in tool configurations, system prompts, and checkpoint availability across the evaluated agents.

\begin{table*}[ht]
\centering
\caption{Locally reproduced evaluation results of baseline models. Each cell reports the reproduced Accuracy (\%) / Average Tool-Call Turns.}
\vspace{2mm}
\label{tab:reproduced_results}
\small
\begin{tabular}{lcccccc}
\toprule
\textbf{Model} & \textbf{MMSearch} & \textbf{FVQA} & \textbf{LiveVQA} & \textbf{BCVL} & \textbf{MMSearch+} & \textbf{IMEB} \\
\midrule
DeepEyes-V2    & 58.20 / 2.11 & 60.64 / 2.84 & 57.98 / 3.68 & 24.48 / 4.30 & 9.46 / 3.91  & 18.00 / 4.71 \\
MMSearch-R1    & 41.80 / 1.40 & 55.85 / 1.28 & 41.84 / 1.43 & 19.05 / 1.71 & 10.14 / 1.76 & 3.33 / 1.91 \\
WebWatcher     & 41.56 / 4.78 & 64.26 / 4.04 & 56.76 / 4.12 & 24.48 / 4.87 & 11.49 / 5.71 & 15.33 / 7.82 \\
VDR-8B         & 38.52 / 11.12& 37.76 / 12.70& 51.37 / 10.18& 39.53 / 11.72& 19.51 / 11.44& 21.18 / 12.34 \\
\bottomrule
\end{tabular}
\end{table*}

\begin{table}[ht]
    \centering
    \caption{Comparison of evaluation environments and tool configurations across different baseline agents. Discrepancies in search APIs, parsing modules, and checkpoint availability significantly impact direct performance comparisons.}
    \label{tab:tool_inconsistency}
    \renewcommand{\arraystretch}{1.2}
    \small
    \setlength{\tabcolsep}{4pt} 
    \begin{tabular}{@{} l c c >{\raggedright\arraybackslash}p{4.8cm} c c c @{}}
        \toprule
        \textbf{Model} & \textbf{Prompt} & \textbf{Ckpt.} & \textbf{Tool Configurations} & \textbf{Temp.} & \textbf{Top-$p$} & \textbf{Turns} \\
        \midrule
        WebWatcher & Yes & Yes & SerpAPI (image \& text) + Jina + Summary model (Qwen2.5-72B) & 0.6 & 0.95 & 12 \\
        \addlinespace
        MMSearch-R1 & Yes & Yes & SerpAPI (image \& text) + Jina + Summary model (Qwen3-32B) & 0.0 & 1.0 & 3 \\
        \addlinespace
        DeepEyes-V2 & Yes & Yes & Code, SerpAPI (image \& text); returns snippets, titles, and links only, without full-page crawling. & 0.0 & 1.0 & 10 \\
        \addlinespace
        VDR & Yes & 8B-SFT & Code, Unspecified commercial search engine + Jina + Summary model (Qwen3-VL-30B-A3B) & 0.6 & 0.95 & 50 \\
        \bottomrule
    \end{tabular}
\end{table}
\begin{table}[ht]
    \centering
    \caption{Evaluation hyperparameters and environment configurations.}
    \label{tab:eval_hyperparameters}
    \renewcommand{\arraystretch}{1.1}
    \small
    \begin{tabular}{@{}lc@{}}
        \toprule
        \textbf{Hyperparameter / Configuration} & \textbf{Value} \\
        \midrule
        \multicolumn{2}{@{}l}{\textit{Generation \& Tokenization}} \\
        Evaluation temperature & 0.0 \\
        Evaluation top-$p$ & 1.0 \\
        Max sequence length & 38,000 tokens \\
        Max visual tokens per image & 1,200 \\
        Max images per rollout & 50 \\
        \midrule
        \multicolumn{2}{@{}l}{\textit{Agentic Constraints}} \\
        Max model turns & 19 \\
        Max tool calls & 18 \\
        \midrule
        \multicolumn{2}{@{}l}{\textit{Search Backends}} \\
        Text search source & SerpAPI (Google Search) \\
        Image search source & SerpAPI (Google Reverse Image Search) \\
        \bottomrule
    \end{tabular}
\end{table}
\paragraph{Inconsistencies in tool environments.}
Although most frameworks adopt SerpAPI for Google text and image searches, their downstream processing pipelines diverge significantly. For instance, while WebWatcher, MMSearch-R1, and VDR all utilize Jina to extract web content, they rely on distinct summary models (Qwen2.5-72B, Qwen3-32B, and Qwen3-VL-30B-A3B respectively) to distill the retrieved text. Conversely, DeepEyes-V2 eschews full-page crawling entirely, restricting its observations to search result snippets, titles, and URLs. These architectural disparities inevitably introduce performance variations that stem from the external tool environment rather than the intrinsic reasoning capabilities of the multimodal agents.

\paragraph{Variations in system prompts and model availability.}
Beyond tool configurations, project-specific system prompts and limited checkpoint availability further complicate the evaluation. Each baseline relies on highly customized system prompts tailored to its specific action schema and reasoning paradigm, making a standardized ``apples-to-apples'' deployment challenging. More critically, incomplete open-source releases hinder fair scale-to-scale comparisons. For example, VDR releases only an 8B supervised fine-tuning checkpoint, withholding the 30B model utilized in its primary evaluations. Consequently, assessing these baselines requires navigating a fragmented landscape of proprietary toolchains and missing model weights, which fundamentally limits the comparability of the reported metrics.

\paragraph{Evaluation hyperparameters.}
To ensure a fair comparison that accurately captures the intended capabilities of existing methods, we evaluate all baseline models using their officially provided system prompts and native hyperparameter settings. Table~\ref{tab:reproduced_results} presents the locally reproduced results of these baselines under their respective original configurations. Conversely, for the evaluation of our proposed model, we establish a standardized and transparent protocol by strictly adhering to the settings detailed in Table~\ref{tab:eval_hyperparameters}. Specifically, we deploy our agent in a zero-shot setting, limiting the interaction to a maximum of 19 turns and 18 tool invocations per query to rigorously measure operational efficiency. During generation, we set the temperature to 0.0 and top-$p$ to 1.0 to encourage deterministic reasoning. Furthermore, we cap the maximum number of visual tokens per image at 1,200, which optimizes memory utilization while preserving sufficient spatial resolution for fine-grained multimodal grounding.

\section{Training Details}
\label{appendix:training}

\paragraph{Implementation Details.}
The training of HyperEyes proceeds in two stages and is run {independently}
for the 30B and 235B model scales, sharing the same data recipes throughout.
In the supervised fine-tuning (SFT) stage, we perform LoRA-based fine-tuning
on the same $30{,}000$ curated trajectories described in
Sec.~\ref{subsec:mt_sft}, with
{Qwen3-VL-30B-A3B-Instruct} and
{Qwen3-VL-235B-A22B-Instruct}~\citep{bai2025qwen3} as the respective
backbones, yielding {HyperEyes-30B-SFT} and
{HyperEyes-235B-SFT}. Each SFT run is conducted on
{$8$ nodes ($64$ NVIDIA H20-141\,GB GPUs)} and converges in
approximately ${10}$\,h for the 30B variant and
${20}$\,h for the 235B variant.

In the reinforcement learning (RL) stage, both variants are further
optimized with GRPO under the proposed TRACE reward on
medium-difficulty subsets curated from the parallel QA corpus
(${6{,}056}$ samples for the 30B variant and
${9{,}337}$ samples for the 235B variant). Specifically, we
configure the TRACE reward with a format penalty of $-0.5$, a
redundancy tolerance factor $\gamma = 1.5$, and a constant redundancy
penalty $\lambda_{red} = -0.1$. The intra-group reward interpolation
bounds are empirically set to $r_{\min}^{+} = 0.05$ and
$r_{\max}^{+} = 0.20$ for strictly efficient trajectories, and
$r_{\min}^{-} = -0.10$ and $r_{\max}^{-} = -0.02$ for valid but
redundant ones. For each prompt, $8$ rollouts are sampled per update.

The two variants then diverge in their RL recipes:
{HyperEyes-235B (RL)} is trained with the TRACE reward
{alone} and, once converged, is reused as the frozen teacher in the
next stage. {HyperEyes-30B (RL)} additionally enables
{On-Policy Distillation} with distillation strength
$\lambda_{kd} = 0.05$, in which this 235B teacher provides token-level
supervision on failed student rollouts in synchrony with GRPO updates. On the same $8$-node ($64$ GPU)
cluster, the RL stage takes approximately ${48}$\,h for
HyperEyes-30B and ${72}$\,h for HyperEyes-235B.

\subsection{Computing Infrastructure}
\label{sec:infra}

\paragraph{Hardware.}
All HyperEyes training runs are conducted on $8$ nodes, each equipped with
$8\times$ NVIDIA H20 ($141$\,GB) GPUs interconnected via NVLink within a
node. The on-policy distillation (OPD) teacher (Sec.~\ref{sec:opd}) is hosted
on a dedicated inference cluster and accessed through an HTTP endpoint, so
that the policy model and the teacher model do not contend for the same
GPUs.

\paragraph{Software \& Framework.}

The two stages of HyperEyes use different open-source training stacks,
each chosen to match the workload of that stage. The SFT stage
(Sec.\ref{sec:sft}) is built on top of {ms-swift}~\citep{zhao2024swift}. The {RL stage} (Sec.\ref{sec:rl}) is built on
top of {Relax}~\citep{relax2026}, an asynchronous reinforcement
learning framework that combines Megatron-LM tensor / expert parallelism on
the trainer side with {SGLang}~\citep{zheng2024sglang} as the
rollout / inference backend. Both stages run on PyTorch~$2.9.1$,
CUDA~$12.9$, and SGLang~$0.5.9$.

\subsection{Supervised Fine-Tuning Stage}
\label{sec:sft}

\paragraph{Backbone \& initialization.}

The SFT stage is run independently for the two model scales. The 30B
variant starts from the publicly released
{Qwen3-VL-30B-A3B-Instruct} checkpoint and produces
{HyperEyes-30B-SFT}; the 235B variant starts from
{Qwen3-VL-235B-A22B-Instruct} and produces
{HyperEyes-235B-SFT}. Both checkpoints serve as the initialization
for their respective RL stages in Sec.\ref{sec:rl}. The two variants share
the same \textasciitilde 30k SFT corpus and identical LoRA / optimization recipe
(Table~\ref{tab:sft_hyperparameters}); only the Megatron parallelism
configuration is adjusted to accommodate the larger backbone.

\subsection{Reinforcement Learning Stage}
\label{sec:rl}
The RL stage produces two variants from a shared GRPO\,+\,TRACE pipeline
that differ in (i) the underlying backbone (30B vs.\ 235B) and (ii) whether
On-Policy Distillation is enabled. {HyperEyes-235B (RL)} is trained
with TRACE only and additionally serves as the OPD teacher for the 30B
variant; {HyperEyes-30B (RL)} adds OPD on top of TRACE
(Sec.\ref{sec:opd}).

\paragraph{Rollout \& generation.}
Rollouts are generated by SGLang with sampling temperature $0.8$ and
top-$p$ $0.95$. The maximum prompt length is $15{,}000$ tokens and the
maximum response length per turn is $1{,}524$ tokens. Each image consumes
at most $1{,}200$ visual tokens, and each rollout may attach up to $50$
images. We use fault-tolerant rollout with shuffling and load balancing.
These generation settings are shared by both 30B and 235B variants.

\paragraph{Parallelism.}
The trainer-side parallelism is adjusted per scale to fit the model in
memory while preserving high MFU. Both variants use Sequence Parallel and
set Expert TP\,$=1$, Context Parallel\,$=1$, with activation recomputation
in uniform mode and dynamic batching enabled. Tensor Parallel and Expert
Parallel sizes differ between scales; full settings are listed in
Table~\ref{tab:rl_hyperparameters}.

\subsection{Search Backends}
\label{sec:search}

To bridge multimodal grounding and open-world knowledge retrieval, the HyperEyes agent is equipped with two complementary tools, both implemented under a unified asynchronous search adapter that supports parallel batched queries within a single turn (multiple queries provided as a list of text strings). These backends are shared across the 30B and 235B variants.

\begin{itemize}
    \item \textbf{Text-search backend.} A real, web-scale retrieval service backed 
    by SerpAPI, a third-party wrapper around the Google Web Search engine. 
    \item \textbf{Image-search backend.} Reverse image search powered by 
    SerpAPI's Google Reverse Image Search endpoint. The agent may operate either on a whole 
    image referenced by \texttt{image\_id}, or on user-specified normalized crop 
    regions $[x_1, y_1, x_2, y_2]$.
\end{itemize}

\paragraph{Tool-call budget.}
Each rollout is capped at \texttt{MAX\_TOOL\_CALLS\_TURN}~$=8$ tool
invocations and \texttt{MAX\_ITERATIONS}~$=9$ model turns. Concurrency to
the retrieval service is throttled to $64$ in-flight requests.

\subsection{On-Policy Distillation Teacher}
\label{sec:opd}

\paragraph{Teacher backend.}

The OPD teacher is exposed as an SGLang \texttt{/generate} HTTP service
(\texttt{opd-type=sglang},
\texttt{rm-url=http://<teacher-host>:30010/generate}). Decoupling the
teacher onto its own SGLang cluster allows us to use a substantially
larger and already RL-aligned sibling model as the teacher without
inflating training-side memory. Concretely, the teacher in our
experiments is {HyperEyes-235B (RL)}, i.e., the converged
TRACE-trained 235B variant from Sec.\ref{sec:rl}. To prevent the teacher
from becoming a bottleneck for asynchronous rollouts, we enforce a
per-actor connector limit of $32$ concurrent requests and a $500$-second
timeout.

\subsection{Hyperparameter Summary}
\label{sec:hp_summary}

The full set of hyperparameters used for the RL stage of both variants is
listed in Table~\ref{tab:rl_hyperparameters}, and the SFT-stage
hyperparameters are listed in Table~\ref{tab:sft_hyperparameters}. Where a
single value is shown, it applies to both 30B and 235B; otherwise the two
columns report the per-variant setting.

\begin{table}[!ht]
\centering
\caption{Hyperparameters for the SFT stage of HyperEyes (LoRA fine-tuning
of Qwen3-VL backbones). Single-value rows apply to both variants;
two-value rows report per-variant settings.}
\vspace{2mm}
\small
\begin{tabular}{lcc}
\toprule
\textbf{Hyperparameter} & \textbf{HyperEyes-30B} & \textbf{HyperEyes-235B} \\
\midrule
\multicolumn{3}{l}{\textit{Backbone}} \\
Pretrained checkpoint           & Qwen3-VL-30B-A3B-Instruct & Qwen3-VL-235B-A22B-Instruct \\
\midrule
\multicolumn{3}{l}{\textit{Optimization \& Training}} \\
Optimizer                       & \multicolumn{2}{c}{Adam ($\beta_1{=}0.9,\ \beta_2{=}0.999$)} \\
Peak learning rate              & \multicolumn{2}{c}{$1\times 10^{-4}$} \\
Min learning rate               & \multicolumn{2}{c}{$1\times 10^{-5}$} \\
LR schedule                     & \multicolumn{2}{c}{cosine decay} \\
Warmup fraction                 & \multicolumn{2}{c}{0.05} \\
Cross-entropy loss fusion       & \multicolumn{2}{c}{enabled} \\
Total epochs                    & \multicolumn{2}{c}{1} \\
\midrule
\multicolumn{3}{l}{\textit{Adaptation Strategy (LoRA)}} \\
Train type                      & \multicolumn{2}{c}{LoRA} \\
LoRA rank $r$                   & \multicolumn{2}{c}{8} \\
LoRA $\alpha$                   & \multicolumn{2}{c}{32} \\
Effective merge scale $\alpha/r$ & \multicolumn{2}{c}{4} \\
Target modules                  & \multicolumn{2}{c}{all-linear} \\
\midrule
\multicolumn{3}{l}{\textit{MoE-Specific}} \\
Grouped GEMM                    & \multicolumn{2}{c}{enabled} \\
Permute fusion                  & \multicolumn{2}{c}{enabled} \\
Shared-expert overlap           & \multicolumn{2}{c}{disabled} \\
Aux load-balance loss coef.     & \multicolumn{2}{c}{$1\times 10^{-3}$} \\
\midrule
\multicolumn{3}{l}{\textit{Batching \& Sequence}} \\
Micro-batch size (per GPU)      & \multicolumn{2}{c}{1} \\
Global batch size               & \multicolumn{2}{c}{64} \\
Max sequence length             & \multicolumn{2}{c}{32{,}000 tokens} \\
Max visual tokens per image     & \multicolumn{2}{c}{1{,}200} \\
\midrule
\multicolumn{3}{l}{\textit{Data}} \\
SFT dialogues                   & \multicolumn{2}{c}{$\sim$30K (shared)} \\
Action schema                   & \multicolumn{2}{c}{\texttt{<reason>/<tool\_call>/<answer>}} \\
Data-loader workers / rank      & \multicolumn{2}{c}{8} \\
\midrule
\multicolumn{3}{l}{\textit{Megatron Parallelism (Trainer)}} \\
Tensor parallel size            & \multicolumn{2}{c}{2} \\
Expert parallel size            & 2 & 64 \\
Expert tensor parallel size     & \multicolumn{2}{c}{1} \\
Sequence parallel               & \multicolumn{2}{c}{enabled} \\
Recompute (granularity/method/layers) & \multicolumn{2}{c}{full / uniform / 1} \\
\midrule
\multicolumn{3}{l}{\textit{Infrastructure}} \\
Nodes / GPUs per node           & \multicolumn{2}{c}{8 / 8 (NVIDIA H20 141\,GB)} \\
\bottomrule
\end{tabular}
\label{tab:sft_hyperparameters}
\end{table}

\begin{table}[!ht]
\centering
\caption{Hyperparameters for the RL stage of HyperEyes-30B (RL) and
HyperEyes-235B (RL). Single-value rows apply to both variants; two-value
rows report per-variant settings. }
\vspace{2mm}
\small
\begin{tabular}{lcc}
\toprule
\textbf{Hyperparameter} & \textbf{HyperEyes-30B (RL)} & \textbf{HyperEyes-235B (RL)} \\
\midrule
\multicolumn{3}{l}{\textit{Framework}} \\
Rollout backend                 & \multicolumn{2}{c}{SGLang~\citep{zheng2024sglang}} \\
Training backend                & \multicolumn{2}{c}{Megatron-LM~\citep{shoeybi2019megatron}} \\
\midrule
\multicolumn{3}{l}{\textit{Optimization \& Training}} \\
Learning rate (LR)              & \multicolumn{2}{c}{$1\times 10^{-6}$} \\
Optimizer                       & \multicolumn{2}{c}{Adam ($\beta_1{=}0.9,\ \beta_2{=}0.98$)} \\
LR schedule                     & \multicolumn{2}{c}{constant} \\
Weight decay                    & \multicolumn{2}{c}{0.1} \\
\midrule
\multicolumn{3}{l}{\textit{Data \& Batching}} \\
RL training samples             & 6{,}056 & 9{,}337 \\
Rollout batch size (prompts)    & \multicolumn{2}{c}{32}  \\
Group size $N$ (samples/prompt) & \multicolumn{2}{c}{8} \\
Effective rollout samples       & \multicolumn{2}{c}{256}  \\
\midrule
\multicolumn{3}{l}{\textit{RL Algorithm (GRPO + Dual-Clip + TIS)}} \\
Clip $\epsilon_{\text{low}}$    & \multicolumn{2}{c}{0.20} \\
Clip $\epsilon_{\text{high}}$   & \multicolumn{2}{c}{0.28} \\
Dual-clip $c$                   & \multicolumn{2}{c}{3.0} \\
KL-loss coef.\ (GRPO branch)    & \multicolumn{2}{c}{0.0} \\
Entropy coef.                   & \multicolumn{2}{c}{0.0} \\
TIS                             & \multicolumn{2}{c}{enabled} \\
\midrule
\multicolumn{3}{l}{\textit{Generation \& Tokenization}} \\
Rollout temperature / top-$p$   & \multicolumn{2}{c}{0.8 / 0.95} \\
Eval temperature / top-$p$      & \multicolumn{2}{c}{1.0 / 0.7} \\
Max visual tokens per image     & \multicolumn{2}{c}{1{,}200} \\
Max images per rollout          & \multicolumn{2}{c}{50} \\
Max tool calls / max turns      & \multicolumn{2}{c}{8 / 9} \\
\midrule
\multicolumn{3}{l}{\textit{Search Backends}} \\
Text-search sources             & \multicolumn{2}{c}{SerpAPI (Google Search)} \\
Image-search sources            & \multicolumn{2}{c}{SerpAPI (Google Reverse Image Search)} \\
Concurrent retrieval requests   & \multicolumn{2}{c}{64} \\
\midrule
\multicolumn{3}{l}{\textit{On-Policy Distillation (OPD)~~\small\textcolor{gray}{[applied to HyperEyes-30B only;
HyperEyes-235B (RL) serves as the frozen teacher]}}} \\
Teacher model                   & HyperEyes-235B (RL) & --- \\
Teacher backend                 & SGLang HTTP & --- \\
OPD KL coef.\ $\beta_{\text{OPD}}$ & 0.05 & --- \\
Teacher timeout                 & 500\,s & --- \\
Teacher connector limit         & 32 & --- \\
\midrule
\multicolumn{3}{l}{\textit{Megatron Parallelism (Trainer)}} \\
Tensor parallel size            & \multicolumn{2}{c}{4} \\
Pipeline parallel size          & \multicolumn{2}{c}{8} \\
Expert parallel size            & \multicolumn{2}{c}{8} \\
Expert tensor parallel size     & \multicolumn{2}{c}{1} \\
Context parallel size           & \multicolumn{2}{c}{1} \\
Sequence parallel               & \multicolumn{2}{c}{enabled} \\
Recompute (granularity/method/layers) & \multicolumn{2}{c}{full / uniform / 1} \\
\midrule
\multicolumn{3}{l}{\textit{SGLang Inference (Rollout)}} \\
Tensor parallel size            & 4 & 8 \\
Max running requests            & \multicolumn{2}{c}{64} \\
Chunked prefill                 & \multicolumn{2}{c}{enabled} \\
\midrule
\multicolumn{3}{l}{\textit{Infrastructure \& Scheduling}} \\
Nodes / GPUs per node           & \multicolumn{2}{c}{8 / 8 (NVIDIA H20 141\,GB)} \\
\bottomrule
\end{tabular}
\label{tab:rl_hyperparameters}
\end{table}

\subsection{Robustness to Random Seed}
\label{app:seed_variance}

\paragraph{Protocol.}
All numbers reported in the main paper (Table~\ref{tab:main_results})
are obtained with a fixed random seed (\texttt{seed=42}) for both
training and rollout sampling. To assess the stability of our
post-training recipe, we additionally re-run the entire reinforcement
learning stage of the smaller HyperEyes-30B (RL) variant
under $N = 3$ independent random seeds (seeds $\in \{42,\ 1234,\ 2026\}$), keeping the SFT checkpoint,
training data, hyperparameters, and infrastructure identical across
runs. For each seed, we evaluate the final RL checkpoint on all six
benchmarks using the same protocol described in
Sec.~\ref{subsub:setup}, and report the mean $\pm$ standard
deviation of accuracy (\%) and average tool turns. We focus on the
30B variant due to compute constraints; the 235B variant is reported
with a single seed throughout.

\paragraph{Aggregate results.}
Table~\ref{tab:seed_summary} summarizes the per-benchmark mean $\pm$
std across the three seeds, alongside the single-seed numbers
reported in the main paper. The standard deviation of accuracy is at
most ${1.22}$ points across all six benchmarks (achieved
on MMSearch) and only ${0.19}$ on the six-benchmark
average; the standard deviation of the average tool turns is at most
${0.10}$ per-benchmark and only ${0.02}$ on
average. Both the accuracy and the efficiency claims of HyperEyes are
therefore stable under different random initializations of the RL
stage. 

\begin{table}[!ht]
\centering
\caption{Robustness of HyperEyes-30B (RL) to the RL random seed.
Mean $\pm$ std across $N = 3$ independent RL runs from the same SFT
checkpoint (seeds $\in \{42, 1234, 2026\}$). ``Reported'' denotes the
single-seed number in Table~\ref{tab:main_results}
(\texttt{seed=42}). \emph{Acc} is benchmark accuracy (\%);
\emph{Turns} is the average number of tool calls per query.}
\vspace{2mm}
\label{tab:seed_summary}
\small
\setlength{\tabcolsep}{5pt}
\begin{tabular}{lcccc}
\toprule
& \multicolumn{2}{c}{\textbf{Reported (seed=42)}}
& \multicolumn{2}{c}{\textbf{Mean $\pm$ Std (3 seeds)}} \\
\cmidrule(lr){2-3}\cmidrule(lr){4-5}
\textbf{Benchmark} & Acc & Turns & Acc & Turns \\
\midrule
MMSearch     & 86.9 & 1.6 & $86.6 \pm 1.22$ & $1.57 \pm 0.06$ \\
FVQA         & 79.3 & 1.7 & $78.7 \pm 0.55$ & $1.73 \pm 0.06$ \\
LiveVQA      & 81.6 & 1.7 & $82.4 \pm 0.85$ & $1.73 \pm 0.06$ \\
BCVL         & 57.9 & 2.6 & $57.7 \pm 0.35$ & $2.53 \pm 0.06$ \\
MMSearch+    & 31.5 & 2.3 & $31.2 \pm 0.52$ & $2.40 \pm 0.10$ \\
IMEB         & 46.7 & 3.1 & $46.8 \pm 0.17$ & $3.10 \pm 0.10$ \\
\midrule
\textbf{Average} & \textbf{64.0} & \textbf{2.2} &
$\boldsymbol{63.9 \pm 0.19}$ & $\boldsymbol{2.18 \pm 0.02}$ \\
\bottomrule
\end{tabular}
\end{table}

\paragraph{Per-seed breakdown.}
For full transparency, Table~\ref{tab:seed_full} reports the
per-seed accuracy and average turns on every benchmark. The
six-benchmark average accuracy stays within a $0.4$-point band
($63.7$--$64.1$) and the average tool turns stay within a $0.04$
band ($2.17$--$2.20$) across all three seeds, confirming that the
reported numbers in Table~\ref{tab:main_results} are not the result
of a favorable seed.

\begin{table}[!ht]
\centering
\caption{Per-seed evaluation of HyperEyes-30B (RL). Each cell
reports accuracy (\%)/average tool turns.}
\vspace{2mm}
\label{tab:seed_full}
\small
\setlength{\tabcolsep}{5pt}
\begin{tabular}{lcccc}
\toprule
\textbf{Benchmark}
& \textbf{Seed 42}
& \textbf{Seed 1234}
& \textbf{Seed 2026}
& \textbf{Mean $\pm$ Std} \\
\midrule
MMSearch  & 86.9 / 1.6 & 85.3 / 1.6 & 87.7 / 1.5 & $86.6 \pm 1.22$ / $1.57 \pm 0.06$ \\
FVQA      & 79.3 / 1.7 & 78.7 / 1.7 & 78.2 / 1.8 & $78.7 \pm 0.55$ / $1.73 \pm 0.06$ \\
LiveVQA   & 81.6 / 1.7 & 83.3 / 1.7 & 82.4 / 1.8 & $82.4 \pm 0.85$ / $1.73 \pm 0.06$ \\
BCVL      & 57.9 / 2.6 & 57.3 / 2.5 & 57.9 / 2.5 & $57.7 \pm 0.35$ / $2.53 \pm 0.06$ \\
MMSearch+ & 31.5 / 2.3 & 30.6 / 2.5 & 31.5 / 2.4 & $31.2 \pm 0.52$ / $2.40 \pm 0.10$ \\
IMEB      & 46.7 / 3.1 & 47.0 / 3.0 & 46.7 / 3.2 & $46.8 \pm 0.17$ / $3.10 \pm 0.10$ \\
\midrule
\textbf{Average}
          & \textbf{64.0 / 2.17}
          & \textbf{63.7 / 2.17}
          & \textbf{64.1 / 2.20}
          & $\boldsymbol{63.9 \pm 0.19}$ / $\boldsymbol{2.18 \pm 0.02}$ \\
\bottomrule
\end{tabular}
\end{table}

\section{Further Analysis}
\subsection{More Tool Calls Do Not Imply Higher Accuracy}
\label{appendix:more_calls}

A central design choice in HyperEyes, {progressive rejection sampling} (Sec.~\ref{subsec:mt_sft}), stems from a counterintuitive empirical observation: {blindly increasing the number of tool calls fails to improve, and frequently degrades, final-answer accuracy}. We demonstrate the universality of this phenomenon by evaluating the Qwen3-VL-235B backbone on two representative benchmarks: FVQA for shallow-hop visual question answering, and BCVL for complex multi-hop reasoning. We test five tool-call budgets, comprising fixed limits of \{2, 4, 6, 8\} alongside an unconstrained setting where the model autonomously terminates within an eight-call upper bound. As Table~\ref{tab:budget_ablation} shows, accuracy exhibits a {non-monotonic} relationship with the budget across both benchmarks. On FVQA, accuracy peaks at the smallest budget of 2 calls (66.6) and steadily declines under forced additional calls. Conversely, BCVL performance peaks at 4 calls (36.1) before degrading. Concurrently, the average number of tool-call turns scales roughly linearly with the imposed budget. This indicates that additional latency and token costs fail to yield accuracy gains, and instead induce measurable performance drops in certain regimes. This consistent pattern across shallow and deep tasks confirms that over-retrieval constitutes a general failure mode rather than a benchmark-specific artifact.

\begin{table}[!ht]
\centering
\caption{Effect of tool-call budget on Qwen3-VL-235B across FVQA and BCVL. Each cell reports \textit{Accuracy / Average Turns}. Best accuracy per row is in \textbf{bold}.}
\vspace{2mm}
\label{tab:budget_ablation}
\begin{tabular}{lccccc}
\toprule
\textbf{Budget} & 2 calls & 4 calls & 6 calls & 8 calls & Auto calls \\
\midrule
\textbf{FVQA } & \textbf{66.5} / 1.24 & 65.4 / 1.79 & 66.0 / 1.87 & 64.9 / 1.96 & 64.4 / 1.36 \\
\textbf{BCVL} & 31.3 / 1.24 & \textbf{36.1} / 2.29 & 36.1 / 2.83 & 34.7 / 3.39 & 34.0 / 2.03 \\
\bottomrule
\end{tabular}
\end{table}

\subsection{Validity Analysis of the Unified Grounded Search Paradigm}

\begin{figure}[!htbp]
    \centering
    \includegraphics[width=0.95\linewidth]{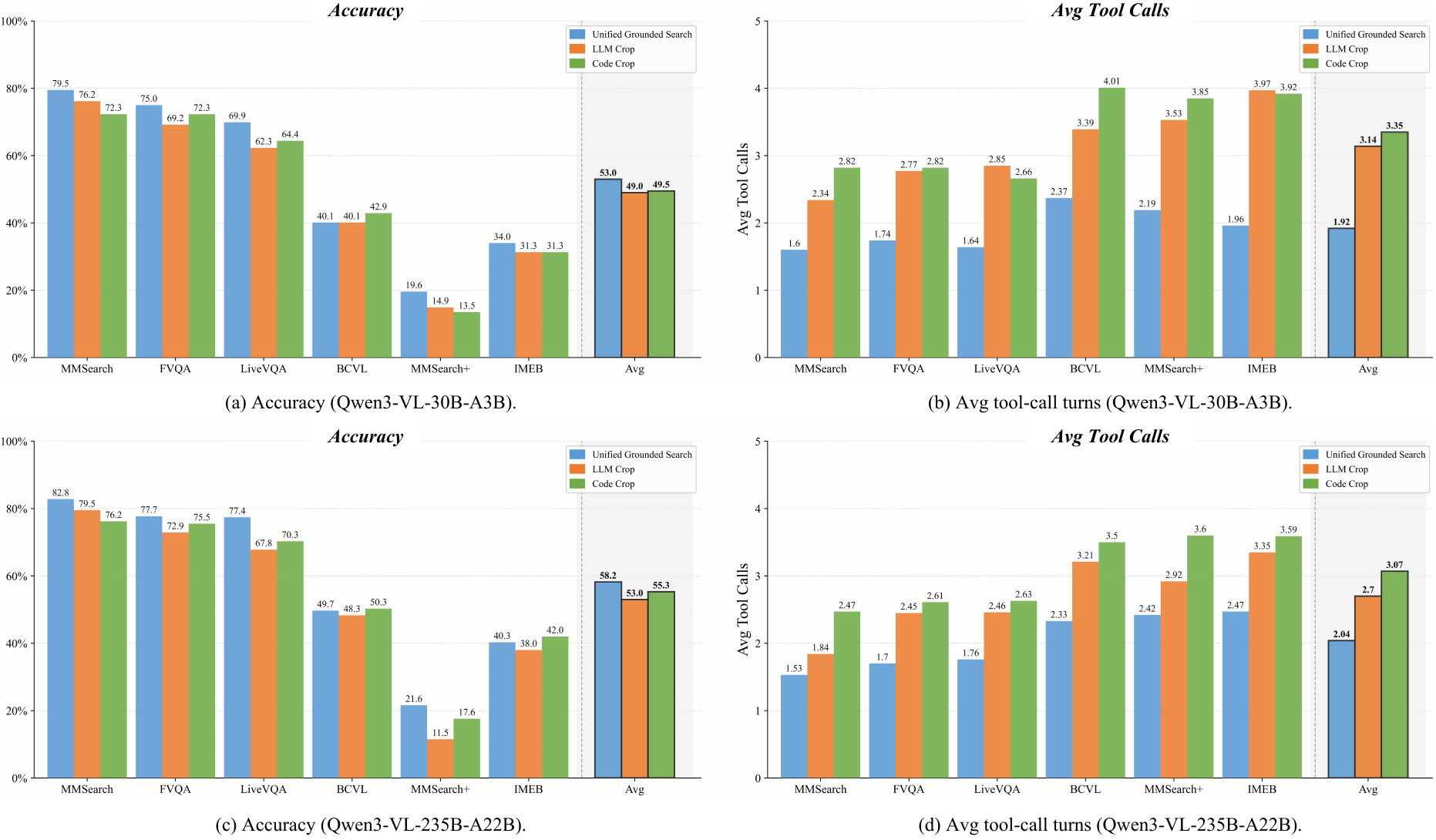}
    \caption{Comparison of three search paradigms under controlled conditions. (a, b) Accuracy and average tool-call turns based on Qwen3-VL-30B-A3B. (c, d) Accuracy and average tool-call turns based on Qwen3-VL-235B-A22B. The "Avg" column denotes the mean across the six benchmarks.}
    \label{fig:search_paradigms}
\end{figure}

\begin{table}[h]
\centering
\caption{Robustness to distractor evidence under shuffled orderings. Each cell reports the mean accuracy across $10$ random shuffles. Both models begin at $100\%$ accuracy under the noise-free regime ($K{=}0$).}
\vspace{2mm}
\label{tab:robustness}
\small
\begin{tabular}{lccccc}
\toprule
\textbf{Method} & $K{=}1$ & $K{=}3$ & $K{=}5$ & $K{=}7$ & $K{=}10$ \\
\midrule
Base (Qwen3-VL-235B)       & 0.890 & 0.890 & 0.896 & 0.873 & 0.883 \\
Ours (HyperEyes-235B SFT)  & \textbf{0.927} & \textbf{0.931} & \textbf{0.915} & \textbf{0.931} & \textbf{0.906} \\
\midrule
$\Delta$Acc & $+3.7\%$ & $+4.1\%$ & $+1.9\%$ & $+5.8\%$ & $+2.3\%$ \\
\bottomrule
\end{tabular}
\end{table}

To verify that the superiority of Unified Grounded Search stems from the paradigm itself rather than confounding factors such as data, backbone, or training scale, we compare three grounding paradigms: {(i) Unified Grounded Search (Ours)}, which jointly emits the bounding boxes and search actions for all target entities in a single decision, collapsing cropping and retrieval into one parallel tool-call round; {(ii) LLM Crop}, where the policy produces a natural-language description of the target region, an external call to Qwen3-VL-235B performs visual grounding to obtain the cropped image; and {(iii) Code Crop}, where the policy generates Python code executed in a sandbox to obtain the cropped image. Their respective system prompts are shown in the Sec.~\ref{Prompt Tempaltes}.

We conduct a strictly controlled comparison across the three approaches commonly adopted by current multimodal agents, holding all other variables constant: the training trajectories are built from the same set of QA pairs synthesized by our Visual Multi-Entity Synthesis pipeline (Sec.~\ref{sec:met_Data} ), with an identical scale of 3k; the Unified Grounded Search and Code Crop trajectories are mutually converted from the same underlying trajectory data, differing only in how grounding is expressed in the action space; all three paradigms are SFT-tuned on Qwen3-VL-30B-A3B and Qwen3-VL-235B-A22B with identical hyperparameters, and evaluated on the same six benchmarks under the same LLM-as-judge and turn budget.

As shown in Fig.\ref{fig:search_paradigms}, across both the 30B and 235B scales and all six benchmarks, Unified Grounded Search consistently achieves the highest average accuracy (58.2 vs. 53.0 / 55.3 on the 235B backbone for LLM Crop and Code Crop, respectively), while also incurring the fewest average tool-call turns (2.04 vs. 2.7 / 3.07 on the 235B backbone). This confirms that Unified Grounded Search delivers higher accuracy with fewer tool calls.

\subsection{Robustness to Distractor Evidence}
\label{sec:robustness}

A practical multimodal search agent must remain reliable when the retrieval 
context contains a mixture of answer-relevant and answer-misleading evidence. 
To quantify how the parallel-amenable cold-start training affects this 
robustness, we conduct a controlled stress test that injects in-domain 
distractor evidence into otherwise-perfect trajectories and measures the 
resulting accuracy degradation.

\paragraph{Experimental setup.}
To isolate the effect of noisy evidence from confounders such as policy 
quality or trajectory length, we adopt a third-party trajectory protocol:

\begin{enumerate}
    \item \textbf{Common reference trajectories.} For each evaluation query, 
    we use a strong external agent (Kimi-K2.5) to generate a successful tool-use 
    trajectory. We then select the $48$ samples on which both the 
    \textit{Base} policy (Qwen3-VL-235B-A22B-Instruct) and \textit{Ours} 
    (HyperEyes-235B SFT) can correctly answer when conditioned on this clean 
    trajectory, ensuring both models start from $100\%$ accuracy in the 
    noise-free regime.
    
    \item \textbf{Distractor synthesis.} For each trajectory, we extract its 
    last-round search query and prompt Gemini-3.0-Flash to generate $10$ 
    in-domain but answer-misleading paraphrased queries. We then issue these 
    distractor queries to SerpAPI and collect their top-$3$ snippets as the 
    distractor evidence pool.
    
    \item \textbf{Noise injection \& shuffling.} We inject $K$ distractor 
    snippets ($K \in \{1, 3, 5, 7, 10\}$) into the last-round retrieval 
    output. To remove any bias from evidence ordering, each (trajectory, $K$) 
    pair is evaluated $10$ times with independently shuffled orderings of the 
    combined evidence list, and we report the mean accuracy across these shuffles.
\end{enumerate}

\rparagraph{Discussion}
As shown in Table~\ref{tab:robustness}, HyperEyes-235B (SFT) outperforms the Qwen3-VL-235B backbone by $1.9\%$--$5.8\%$ accuracy across all tested noise levels, despite both models starting from an identical $100\%$ noise-free accuracy. This indicates that the parallel-amenable SFT corpus instills a sharper sensitivity to answer-relevant evidence, making the policy less susceptible to in-domain distractors. Even under heavy noise at $K{=}10$, where genuine evidence is accompanied by $10\times$ as many distractors, HyperEyes retains $90.6\%$ accuracy, showing a relative drop of less than $10\%$ from the noise-free regime compared to an $11.7\%$ drop for the baseline. These findings demonstrate that the parallel-amenable SFT corpus not only teaches the model to dispatch concurrent searches efficiently, but also implicitly strengthens its evidence-aggregation capability, resulting in a noise-robust policy that generalizes beyond the training distribution.

\subsection{Case Study: DeepEyes-V2 vs. HyperEyes}

\begin{figure}[!h]
    \centering
    \includegraphics[width=1\linewidth]{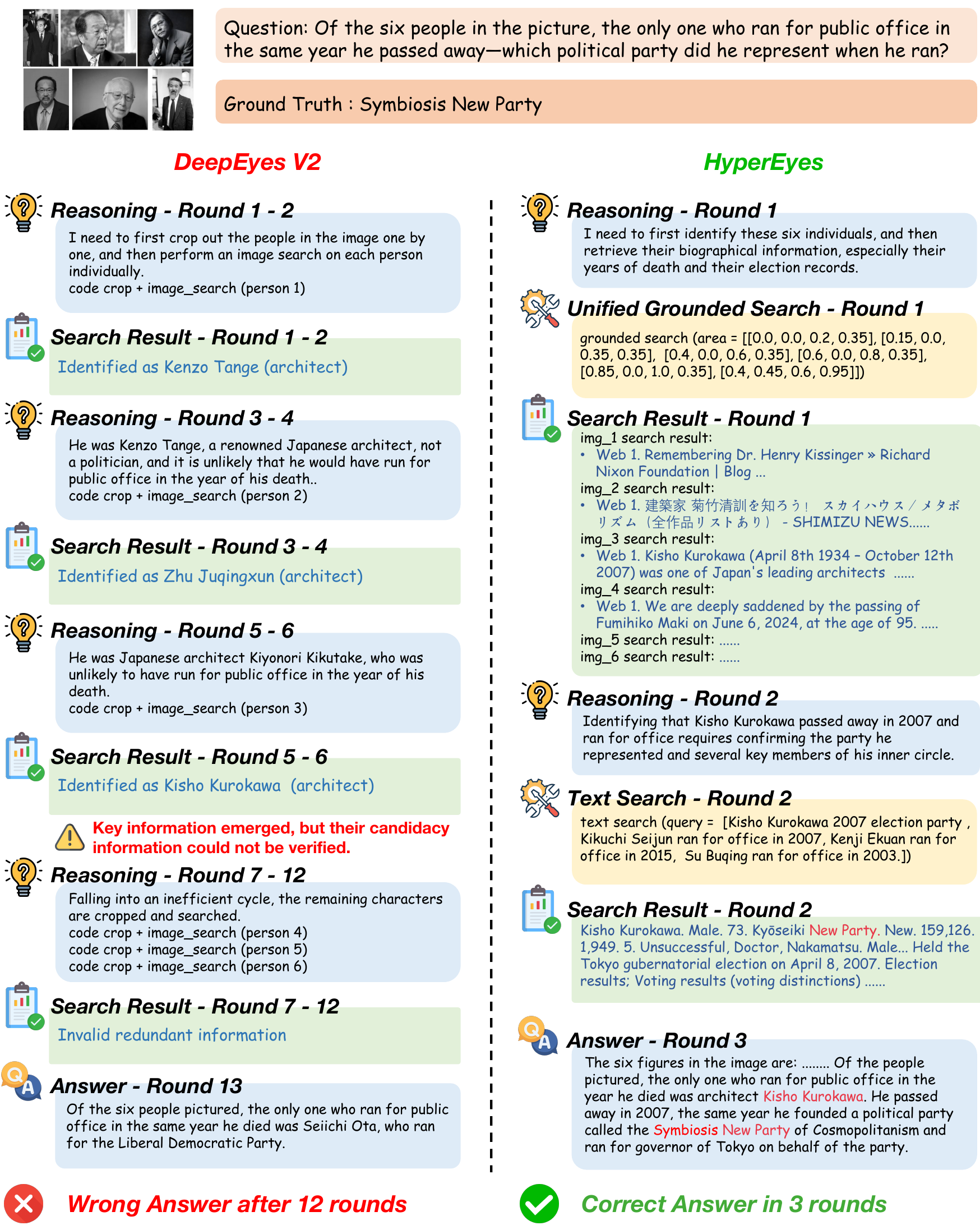}
    \caption{Comparison between the representative serial agent DeepEyes-V2 and our parallel grounded-search framework HyperEyes on a multi-person visual reasoning task. DeepEyes-V2 follows an inefficient crop-then-search pipeline that processes one person at a time, whereas HyperEyes issues a single unified grounded search over all individuals in parallel, followed by a targeted text search to confirm the answer.}
    \label{fig:serial_vs_parallel}
\end{figure}

To intuitively illustrate the efficiency gap between the conventional serial tool-use paradigm and our parallel grounded-search design, we present a head-to-head case study between a representative serial multimodal search agent, {DeepEyes-V2}, and our {HyperEyes} on a multi-entity, knowledge-intensive visual question. As illustrated in Fig~\ref{fig:serial_vs_parallel}, DeepEyes-V2 follows a strictly serial ``crop-then-search'' pipeline: it must isolate and query each person one by one, stretching the trajectory across many redundant rounds (12 in this example) during which intermediate observations accumulate noise and rapidly consume the limited context budget. More critically, this paradigm tightly couples entity identification with attribute verification: even when a key candidate is correctly recognized early on (e.g., Kisho Kurokawa at Round 5--6), DeepEyes-V2 cannot pause to verify his candidacy and is ultimately forced to commit to an incorrect answer (Seiichi Ota / Liberal Democratic Party) based on incomplete evidence.

In contrast, HyperEyes issues a \textit{unified grounded search} that localizes and retrieves biographical evidence for all six individuals in a single round, decoupling identification from downstream reasoning. With identities resolved in parallel, the model devotes its second round to a focused text search on the verified candidate's election record, arriving at the correct answer (Symbiosis New Party) in only 3 rounds. This comparison demonstrates that, relative to a strong serial baseline like DeepEyes-V2, our parallel grounded-search design reduces tool-use rounds by roughly 4× while substantially improving answer accuracy on multi-entity, knowledge-intensive visual questions.

\section{Prompt Templates}
\label{Prompt Tempaltes}

Tables~\ref{fig:prompt_hypereys_system_prompt}, \ref{fig:prompt_llm_crop}, \ref{fig:prompt_code_crop}, \ref{fig:llm_judge_prompt}and \ref{fig:prompt_system_prompt_for_trajectory_construction} detail the prompts utilized for our Unified Grounded Search framework, the comparative baseline grounding variants,evaluation and the trajectory construction pipeline.

\begin{center}
\begin{tcolorbox}[colback=lightbluebg!30!white,colframe=blueframe,breakable,title=HyperEyes System Prompt (Unified Grounded Search)]
\begin{VerbatimWrap}
You are a multimodal intelligent assistant capable of using tools to answer users' questions.

Tool Definitions:
1. image_search: Image search, retrieves web content of similar images
    - `image_id` (string): Image ID, e.g., "img_0", "img_1"
    - `area` (list[list[float]], optional): List of normalized coordinates [[x1,y1,x2,y2], ...], specifying the target region to be searched
        Coordinate range 0.0~1.0 (x corresponds to the width direction, y corresponds to the height direction)

2. text_search: Text search, retrieves relevant web content
    - `input` (list[string]): Query content, supports passing multiple queries at once

Tool Call Format Examples:
- Image search (full image): <tool_call>{"name": "image_search", "arguments": {"image_id": "img_0"}}</tool_call>
- Image search (region): <tool_call>{"name": "image_search", "arguments": {"image_id": "img_0", "area": [[0.1, 0.2, 0.5, 0.8], ... ]}}</tool_call>
- Text search: <tool_call>{"name": "text_search", "arguments": {"input": ["fun places in Shenzhen", "must-eat food in Shenzhen"]}}</tool_call>

Action Guidelines:
- Only one tool can be called at a time.
- Before taking any action, you must enclose your reasoning between `<reason>` and `</reason>`.
- When you need to identify image content, use the image_search tool. If the image background is complex or there are multiple subjects, specify the target region via the `area` parameter to ensure the retrieved region has a single, clear subject.
- When lacking relevant knowledge to answer a question, use the text_search tool, which supports multiple queries in parallel.
- All tool call results will be returned to you placed between <tool_response> and </tool_response>.
- When you are ready to answer the question, please put the final answer between <answer> and </answer>.
\end{VerbatimWrap}
\end{tcolorbox}
\vspace{2mm}
\captionof{table}{System prompt of HyperEyes for unified grounded search, which allows the model to perform region-level grounded retrieval by passing one or multiple normalized bounding boxes within a single image\_search call}
\label{fig:prompt_hypereys_system_prompt}
\end{center}

\begin{center}
\begin{tcolorbox}[colback=lightbluebg!30!white,colframe=blueframe,breakable,title=LLM Crop System Prompt]
\begin{VerbatimWrap}
You are a multimodal intelligent assistant capable of using tools to answer users' questions.

Tool Definitions:

1. crop_image: Image cropping tool for precisely extracting target regions from images
    - `image_id` (string): Source image ID, e.g., "img_0", "img_1"
    - `prompt` (string): Natural language description of the region to crop; use directional terms or subject names, and the system will locate and crop accordingly

2. image_search: Image search, retrieves web content of similar images
    - `image_id` (string): Image ID, e.g., "img_0", "img_1"

3. text_search: Text search, retrieves relevant web content
    - `input` (list[string]): Query content, supports passing multiple queries at once

Tool Call Format Examples:
  - Image crop: <tool_call>{"name": "crop_image", "arguments": {"image_id": "img_0", "prompt": "the dog in the upper left corner"}}</tool_call>
  - Image search: <tool_call>{"name": "image_search", "arguments": {"image_id": ["img_0", "img_1"]}}</tool_call>
  - Text search: <tool_call>{"name": "text_search", "arguments": {"input": ["fun places in Shenzhen", "must-eat food in Shenzhen"]}}</tool_call>

Action Guidelines:

  - Only one tool can be called at a time.
  - Before taking any action, you must enclose your reasoning between `` and ``.
  - When you need to identify image content, use the image_search tool. If the image background is complex or there are multiple subjects, use crop_image first to ensure the retrieved region has a single, clear subject.
  - When lacking relevant knowledge to answer a question, use the text_search tool, which supports multiple queries in parallel.
  - All tool call results will be returned to you placed between <tool_response> and </tool_response>.
  - When you are ready to answer the question, please put the final answer between <answer> and </answer>. 
\end{VerbatimWrap} 
\end{tcolorbox}
\vspace{2mm}
\captionof{table}{System prompt of the LLM-Crop variant, where the model localizes target regions through a language-prompted crop\_image tool before performing image\_search on the cropped output.}
\label{fig:prompt_llm_crop}
\end{center}

\begin{center}
\begin{tcolorbox}[colback=lightbluebg!30!white,colframe=blueframe,breakable,title=Code Crop System Prompt]
\begin{VerbatimWrap}
You are a multimodal intelligent assistant capable of using tools to answer users' questions.

Tool Definitions:
1. python: Python code cropping tool that runs in a Jupyter Notebook environment, used to crop one or more regions from existing images.
    - Input images are preloaded as global variables img_0, img_1, ... (PIL.Image format)
    - Multiple regions from the same image or multiple existing images can be cropped within a single <code> block
    - Use plt.show() to display cropped images; the system will automatically return and register them as new image IDs
    - Code is placed inside <code> tags, wrapped with ```python ... ```
    - Note: All code runs in the same Jupyter Notebook kernel; functions and variables are automatically saved.

2. image_search: Reverse image search to retrieve similar images and contextual information.
    - `image_id` (list[string]): List of target image IDs, e.g., ["img_0", "img_1"]
    - For better accuracy, ensure the image used for search has a single, clear subject; recommend cropping with python first before searching.

3. text_search: Text search to retrieve relevant web content.
    - `input` (list[string]): Query content; supports multiple queries at once for parallel retrieval.

Tool Call Format Examples:
- Python code crop (crop region of interest from image):
<code>
```python
w, h = img_0.size
# Crop the object in the upper-left area of the image
crop_1 = img_0.crop((int(w*0.1), int(h*0.2), int(w*0.5), int(h*0.8)))
plt.imshow(crop_1)
plt.axis('off')
plt.show()

# Crop the object in the lower-right corner of the image
crop_2 = img_0.crop((int(w*0.6), int(h*0.5), int(w*0.95), int(h*0.95)))
plt.imshow(crop_2)
plt.axis('off')
plt.show()
```
</code>
- Image search: <tool_call>{"name": "image_search", "arguments": {"image_id": ["img_1", "img_2"]}}</tool_call>
- Text search: <tool_call>{"name": "text_search", "arguments": {"input": ["fun places in Shenzhen", "must-eat food in Shenzhen"]}}</tool_call>

Action Guidelines:
- Only one tool can be called at a time.
- Before taking any action, you must enclose your reasoning between `<reason>` and `</reason>`.
- When images contain multiple subjects, complex backgrounds, or require local region identification, use python code to crop the image. Each plt.show() generates a new image and registers a new image ID.
- When unable to identify image content, use the image_search tool. For better accuracy, ensure the image used for retrieval has a single, clear subject.
- When lacking relevant knowledge to answer a question, use the text_search tool, which supports multiple text queries in parallel.
- All tool call results will be returned to you placed between <tool_response> and </tool_response>.
- When you are ready to answer the question, please put the final answer between <answer> and </answer>.
\end{VerbatimWrap}
\end{tcolorbox}
\vspace{2mm}
\captionof{table}{System prompt of the Code-Crop variant, in which the model grounds target regions by generating Python cropping code (executed in a Jupyter kernel) before performing image\_search on the produced sub-images.}
\label{fig:prompt_code_crop}
\end{center}

\begin{center}
\begin{tcolorbox}[colback=lightbluebg!30!white,colframe=blueframe,breakable,title=LLM Judge Prompt]
\begin{VerbatimWrap}
Judge whether the following [response] to [question] is correct or not based on the precise and unambiguous [correct_answer] below.

[question]: {question}

[response]: {response}

[correct_answer]: {correct_answer}

Your judgement must be in the following JSON format exactly:
{{
  "extracted_final_answer": "The final exact answer extracted from the [response]. Put 'None' if there is no exact answer.",
  "reasoning": "Explain why the extracted answer is correct or incorrect, focusing only on meaningful differences. Do not attempt to solve the problem.",
  "correct": "yes or no",
  "confidence": 100
}}

An example output:
{{
  "extracted_final_answer": "September 20, 2024",
  "reasoning": "The extracted answer matches the correct answer exactly.",
  "correct": "yes",
  "confidence": 95
}}

Now output the JSON only, no other text:
\end{VerbatimWrap} 
\end{tcolorbox}
\vspace{2mm}
\captionof{table}{Prompt template of the LLM-as-a-judge used for benchmark accuracy evaluation.}
\label{fig:llm_judge_prompt}
\end{center}

\begin{center}
\begin{tcolorbox}[colback=lightbluebg!30!white,colframe=blueframe,breakable,title=System Prompt for Trajectory Construction]
\begin{VerbatimWrap}
You are a multimodal intelligent assistant capable of using tools to answer users' questions.

Tool Definitions:
1. image_search: Image search, retrieves web content of similar images
    - `image_id` (string): Image ID, e.g., "img_0", "img_1"
    - `area` (list[list[float]], optional): List of normalized coordinates [[x1,y1,x2,y2], ...], specifying the target region to be searched
        Coordinate range 0.0~1.0 (x corresponds to the width direction, y corresponds to the height direction)

2. text_search: Text search, retrieves relevant web content
    - `input` (list[string]): Query content, supports passing multiple queries at once

Tool Call Format Examples:
- Image search (full image): <tool_call>{"name": "image_search", "arguments": {"image_id": "img_0"}}</tool_call>
- Image search (region): <tool_call>{"name": "image_search", "arguments": {"image_id": "img_0", "area": [[0.1, 0.2, 0.5, 0.8], ... ]}}</tool_call>
- Text search: <tool_call>{"name": "text_search", "arguments": {"input": ["fun places in Shenzhen", "must-eat food in Shenzhen"]}}</tool_call>

Action Guidelines:
- Only one tool can be called at a time.
- Before taking any action, you must enclose your reasoning between `<reason>` and `</reason>`.
- When you need to identify image content, use the image_search tool. If the image background is complex or there are multiple subjects, specify the target region via the `area` parameter to ensure the retrieved region has a single, clear subject.
- When the question references multiple visual entities (e.g., several people, objects, landmarks, or animals) that all need to be identified, you must localize all of them in a single image_search call by passing every bounding box together in the `area` list, rather than cropping and searching them one by one across multiple turns.
- When lacking relevant knowledge to answer a question, use the text_search tool, which supports multiple queries in parallel. Independent sub-questions (e.g., verifying different attributes about different entities) must be batched into a single text_search call via the `input` list, rather than issued in separate turns.
- Do not issue duplicate queries, near-duplicate paraphrases, or speculative searches whose results you have no concrete plan to use. 
- As soon as the accumulated evidence is sufficient to answer the question, stop calling tools and emit the final answer. Additional "just-to-be-safe" tool calls should be avoided.
- All tool call results will be returned to you placed between <tool_response> and </tool_response>.
- When you are ready to answer the question, please put the final answer between <answer> and </answer>.
\end{VerbatimWrap}
\end{tcolorbox}
\vspace{2mm}
\captionof{table}{System prompt for trajectory construction. It tightens the action guidelines of the unified grounded-search prompt by requiring the model to (i) batch all target regions into one image\_search call, (ii) batch all independent sub-queries into one text\_search call, and (iii) stop calling tools once the evidence is sufficient.}
\label{fig:prompt_system_prompt_for_trajectory_construction}
\end{center}

\end{document}